\documentclass[journal]{IEEEtran}
\usepackage{url}
\usepackage{color}
\usepackage{booktabs}
\usepackage{multirow}
\usepackage{multicol}
\usepackage{graphicx}
\usepackage{subfigure}
\usepackage{algorithm}
\usepackage{algorithmic}
\usepackage{threeparttable}
\usepackage{lineno,amsmath,graphicx}
\usepackage[colorlinks,
            linkcolor=red,
            anchorcolor=blue,
            citecolor=green
            ]{hyperref}

\begin{document}
\title{Deep Semantic Parsing of Freehand Sketches with Homogeneous Transformation, Soft-Weighted Loss, and Staged Learning}

\author{Ying~Zheng,~Hongxun~Yao$^*$,~\IEEEmembership{Member,~IEEE,}~and~Xiaoshuai~Sun% <-this % stops a space
\thanks{The work was supported in part by the National Science Foundation of
China (61772158 and 61702136) and Research Program of Zhejiang Lab (2019KD0AC02 and 2020KD0AA02).}% <-this % stops a space
\thanks{Y. Zheng is with the Artificial Intelligence Research Institute, Zhejiang Lab, Hangzhou 310023, and with the School of Computer Science and Technology, Harbin Institute of Technology, Harbin 150001, China (e-mail: zhengyinghit@outlook.com).}% <-this % stops a space
\thanks{H. Yao is with the School of Computer Science and Technology, Harbin Institute of Technology, Harbin 150001, China (e-mail: h.yao@hit.edu.cn).}% <-this % stops a space
\thanks{X. Sun is with the School of Informatics, Xiamen University, Xiamen 361005, China (e-mail: xssun@xmu.edu.cn).}% <-this % stops a space
\thanks{Corresponding author: Hongxun~Yao.}% <-this % stops a space
}

\markboth{IEEE Transactions on Multimedia, Manuscript}%
{Zheng \MakeLowercase{\textit{et al.}}: Bare Demo of IEEEtran.cls for IEEE Journals}

\maketitle

\begin{abstract}
In this paper, we propose a novel deep framework for part-level semantic parsing of freehand sketches, which makes three main contributions that are experimentally shown to have substantial practical merit. First, we propose a homogeneous transformation method to address the problem of domain adaptation. For the task of sketch parsing, there is no available data of labeled freehand sketches that can be directly used for model training. An alternative solution is to learn from datasets of real image parsing, while the domain adaptation is an inevitable problem. Unlike existing methods that utilize the edge maps of real images to approximate freehand sketches, the proposed homogeneous transformation method transforms the data from domains of real images and freehand sketches into a homogeneous space to minimize the semantic gap. Second, we design a soft-weighted loss function as guidance for the training process, which gives attention to both the ambiguous label boundary and class imbalance. Third, we present a staged learning strategy to improve the parsing performance of the trained model, which takes advantage of the shared information and specific characteristic from different sketch categories. Extensive experimental results demonstrate the effectiveness of the above three methods. Specifically, to evaluate the generalization ability of our homogeneous transformation method, additional experiments for the task of sketch-based image retrieval are conducted on the QMUL FG-SBIR dataset. Finally, by integrating the proposed three methods into a unified framework of deep semantic sketch parsing (DeepSSP), we achieve the state-of-the-art on the public SketchParse dataset.
\end{abstract}


\IEEEpeerreviewmaketitle

\section{Introduction}
\label{sec:introduction}
\IEEEPARstart{F}{reehand} sketch analysis is an important research topic in the multimedia community, especially for applications of content-based retrieval \cite{bhattacharjee2018query} \cite{choi2019sketchhelper} and cross-media computing \cite{ma2012sketch} \cite{ji2018cross}. Due to the strong abstract ability to express objects and scenes, the freehand sketch gains great interest from lots of researchers over the past decade. Most of sketch related works focus on the task of sketch-based image \cite{wang2015sketchmm} \cite{zhang2016sketch} \cite{tolias2017asymmetric} and 3D retrieval \cite{shao2011discriminative} \cite{wang2015sketch} \cite{xu2013sketch2scene}, sketch parsing \cite{sun2012free} \cite{schneider2016example} \cite{sarvadevabhatla2017sketchparse} and recognition \cite{li2015free} \cite{zhang2016sketchnet} \cite{yu2017sketch}, conversion between the real image and sketch \cite{chen2009sketch2photo} \cite{chen2013poseshop} \cite{qi2015im2sketch}. In this paper, we aim at exploring the problem of part-level freehand sketch parsing. The in-depth understanding and further solution to this problem can facilitate the development of sketch related applications, such as sketch captioning, drawing assessment, and sketch-based image retrieval.

The existing works of freehand sketch parsing mainly focus on stroke-level labeling, which group strokes or line segments into semantically meaningful object parts \cite{schneider2016example} \cite{li2018universal}. This kind of labeling method is quite different from the semantic parsing of real images, as the former only needs to post the semantic label to each pixel in the sketch stroke, while the latter requires complete labeling of every pixel in the real image. Due to this difference, lots of existing methods for real image parsing cannot be directly applied in the field of stroke-level labeling.

\begin{figure}[t]
    \centering
    \centerline{\includegraphics[width=1.0\linewidth]{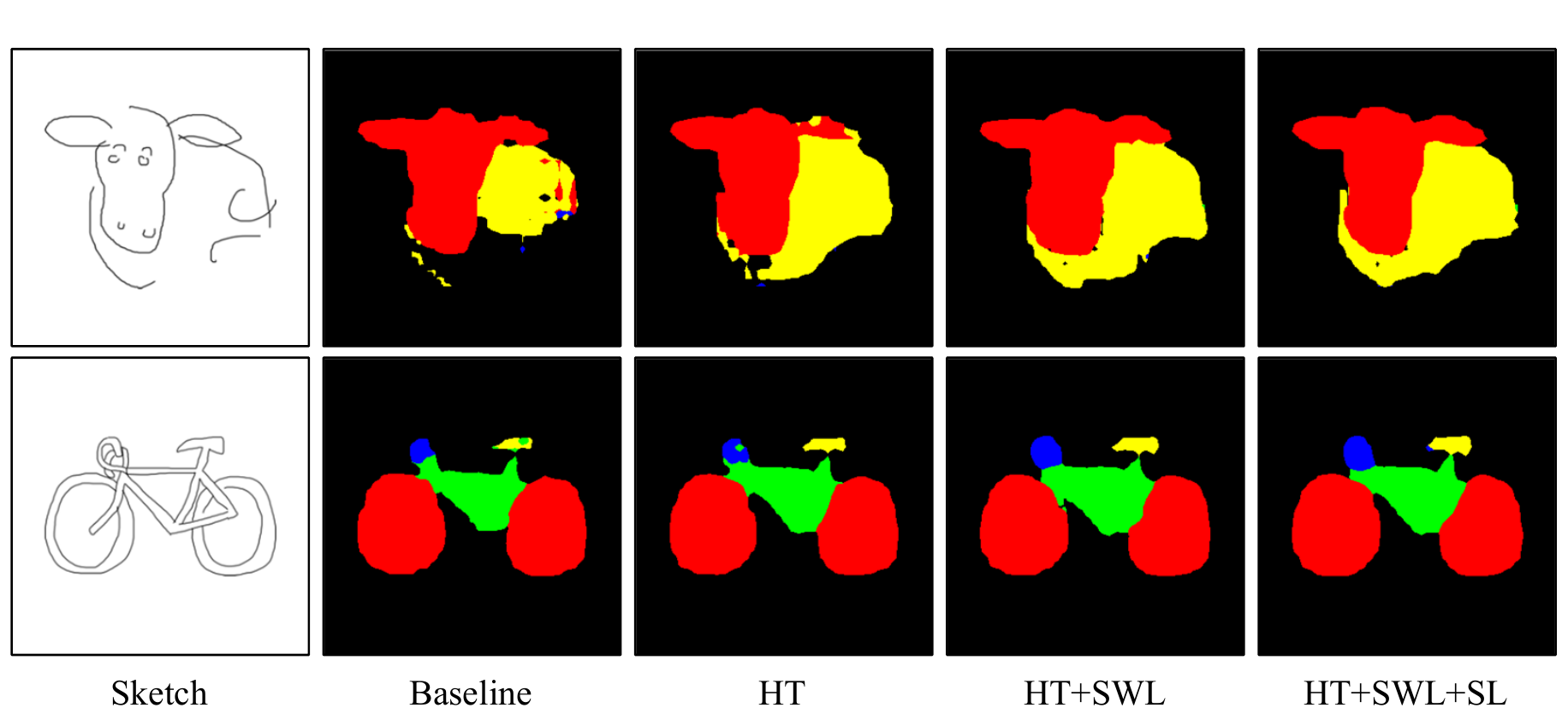}}
    \caption{Comparative results of the baseline and our methods for the freehand sketch parsing. HT: homogeneous transformation, SWL: soft-weighted loss, SL: staged learning. It shows that incorporating these methods generates the best parsing result.}
    \label{fig:contribution}
\end{figure}

However, the part-level sketch parsing does not have this limitation, which also conducts complete pixel-level labeling like the real image parsing, as illustrated in Fig. \ref{fig:contribution}. Intuitively, existing advanced methods for real image parsing can be transferred directly to the task of part-level sketch parsing. However, this would degrade the parsing performance drastically on sketch datasets due
to three challenges: (1) semantic gap between domains of real images and freehand sketches, (2) ambiguous label boundary and class imbalance, (3) information sharing across different sketch categories. Next, we discuss these challenges and present our methods to overcome them in the proposed deep semantic sketch parsing (DeepSSP) framework.

The semantic gap emerges in dealing with different data domains or models. Sarvadevabhatla et al. \cite{sarvadevabhatla2017sketchparse} first propose the task of part-level semantic sketch parsing and release the SketchParse dataset for evaluation. Nevertheless, there is still no available labeled data for end-to-end training of parsing models. A promising solution is to utilize datasets from the area of real image parsing to train the sketch parsing model. However, as the training data comes from the domain of real images, it is inevitable to face the challenge of semantic gap between the domains of real images and freehand sketches. To solve this problem, several existing works directly take the edge maps of real images as a simulate form to the freehand sketch \cite{sarvadevabhatla2017sketchparse} \cite{song2017deep}. Different from them, we propose to transform the edge maps of real images and freehand sketches into a homogeneous space, in which two kinds of data represent the same property. In particular, we define the ``stroke thickness'' as one property of the homogeneous space and convert all edge maps and sketches into 1-pixel thickness. The homogeneous transformation is straightforward, but it makes the training of the sketch parsing model more effective.

The second challenge comes with the inherent nature of the sketch data, which consists of two aspects: ambiguous label boundary and class imbalance. The former is caused by the high abstraction of freehand sketches. The sketch only needs a small number of stroke lines to describe an object and lacks cues of color and texture. It makes the label boundary of adjacent parts ambiguous. The class imbalance indicates the variation in the number of training samples for different classes, which is a common problem in many fields of computer vision, such as image classification \cite{huang2016learning} and object detection \cite{shrivastava2016training}. For the part-level freehand sketch parsing, the number of pixels belonging to each class of the semantic part is dramatically different. Taking the category ``horse'' as an example, the number of pixels belonging to the part class ``torso'' is hundreds of times compared with the class ``tail''. To address these issues, we propose a soft-weighted loss function that acts as more effective supervision for model training.

The third challenge equals to the question ``how to make the best use of the information shared among different sketch categories to learn a better sketch parsing model?''. An alternative solution for the task of sketch parsing is to train a category-specific model for each category due to the label discrepancy. However, this solution largely limits the model generalization capability and generates too many independent models, which leads to inconvenient for training and test. To overcome this challenge, we present a staged learning strategy to make better use of the shared information across categories. At the first stage, the training data of all categories are used to learn the parameters of shared layers under a super branch architecture. At the next stage, we freeze the shared layers and replace the super branch to several category-specific branches. Then, we utilize the training data of each category to train the layers in the corresponding branch. This strategy takes consideration of the information sharing and specific characteristic respectively during these stages, which effectively improve the parsing performance of the trained model.

Extensive experimental results on the SketchParse dataset demonstrate the effectiveness of our methods for part-level freehand sketch parsing. In particular, as a general method for domain adaptation between real images and freehand sketches, we further demonstrate that our homogeneous transformation is also very effective in improving the performance of deep models on the task of fine-grained sketch-based image retrieval (FG-SBIR). Furthermore, we present an erasing-based augmentation method to enhance the training data. After incorporating the proposed methods into the deep semantic sketch parsing (DeepSSP) framework, we achieve the state-of-the-art performance on the SketchParse dataset. To illustrate the contributions of the proposed methods, we show the comparative results in Fig. \ref{fig:contribution}.

The innovation and contributions of this paper are summarized as follows:
\begin{enumerate}
\item We introduce a homogeneous transformation method to reduce the gap between domains of real images and freehand sketches for sketch related tasks, such as the sketch parsing and sketch-based image retrieval.
\item We propose a soft-weighted loss that acts as an effective solution to the problem of ambiguous label boundary and class imbalance.
\item We present a staged learning strategy to further enhance the parsing ability of the trained model for each sketch category.
\item Extensive experimental results demonstrate the practical value of the proposed DeepSSP framework, which outperforms the existing state-of-the-art approaches on the public SketchParse dataset.
\end{enumerate}

The remaining sections are organized as follows. We first briefly review the related work in fields of the semantic image/sketch parsing and domain adaptation in Section \ref{sec:relatedwork}. We give detailed descriptions of the proposed homogeneous transformation, soft-weighted loss and staged learning in Section \ref{sec:methods}. Experimental results, comprehensive analysis, implementation details, and discussions are provided in Section \ref{sec:experiments}. Finally, we articulate our conclusions in Section \ref{sec:conclusion}.

\section{Related Work}
\label{sec:relatedwork}
In this section, we first briefly review two branches of works in the field of semantic image parsing: object-level segmentation and part-level parsing. Then we move on to the representative works of semantic sketch parsing, including the stroke-level labeling and part-level parsing. As the domains of real images and freehand sketches are involved in this paper, we also introduce some related work of domain adaptation.

\subsection{Semantic Image Parsing}
\textbf{Object-level segmentation.} Convolutional neural networks are making significant progress on the task of image classification \cite{krizhevsky2012imagenet}, which shows a promising direction to drive advances in semantic segmentation. The first work exploring the capabilities of existing networks for semantic image segmentation is proposed by Long et al. \cite{long2015fully}. They combine the well-known CNN models for image classification (e.g. AlexNet \cite{krizhevsky2012imagenet}, VGG \cite{simonyan2014very}, and GoogleNet \cite{szegedy2015going}) with fully convolutional networks (FCN) to make dense predictions for every pixel. Following the success of FCN, lots of networks and filters are developed to improve the performance of semantic segmentation, such as DeconvNet \cite{noh2015learning}, U-Net \cite{ronneberger2015u}, DeepLab \cite{chen2018deeplab}, PSPNet \cite{zhao2017pyramid}, and SegNet \cite{badrinarayanan2017segnet}. These methods label each pixel of the real image with the class of its belonged object or region, while do not distinguish the instances from the same class. To output a finer result,  deep frameworks like FCIS \cite{li2017fully} and Mask R-CNN \cite{he2017mask} are designed to separate different instances with the same class label. Kang et al. \cite{kang2018depth} propose to utilize the depth map in a depth-adaptive network for sematic segmentation. Furthermore, recent methods like PGNet \cite{zhang2019pyramid} and SpyGR \cite{li2020spatial} explore the power of capturing global information in graph neural networks for semantic segmentation.

\textbf{Part-level parsing.} Compared to object-level segmentation, part-level parsing focuses on decomposing segmented objects into semantic components. Wang et al. \cite{wang2015joint} propose to jointly solve the problem of object segmentation and part parsing by using a two-stream fully convolutional networks (FCN) and deep learned potentials. Liang et al. \cite{liang2016semantic} design a deep local-global long short-term memory (LG-LSTM) architecture for part-level semantic parsing, which learns features in an end-to-end manner instead of employing separate post-processing steps. To generate high-resolution predictions, Lin et al. \cite{lin2017refinenet} present a generic multi-path refinement network (RefineNet) that exploits features at multiple levels. Beyond these works on general object categories, there also exist some methods specifically designed for human parsing \cite{yamaguchi2012parsing} \cite{dong2014towards} \cite{liang2016clothes}. Liang et al. \cite{liang2015human} integrate different kinds of context like the cross-layer context and cross-super-pixel neighborhood context into a contextualized convolutional neural network (Co-CNN). With the consideration of human body configuration, Gong et al. \cite{gong2017look} propose a self-supervised structure-sensitive learning method and release a new human parsing dataset named ``Look into Person (LIP)''. The biggest difference between these methods and ours is that the starting point of this paper lays in the freehand sketch parsing, which faces several unique challenges as mentioned in Section \ref{sec:introduction}.

\subsection{Semantic Sketch Parsing}
With the strong abstract representation ability of sketches, semantic sketch parsing can be an effective tool to facilitate other applications of computer vision, such as object detection \cite{bhattacharjee2015part} and image retrieval \cite{seddati2017quadruplet}. According to the difference of output, existing works of semantic sketch parsing can be divided into the following two categories.

\begin{figure}[t]
    \centering
    \centerline{\includegraphics[width=1.0\linewidth]{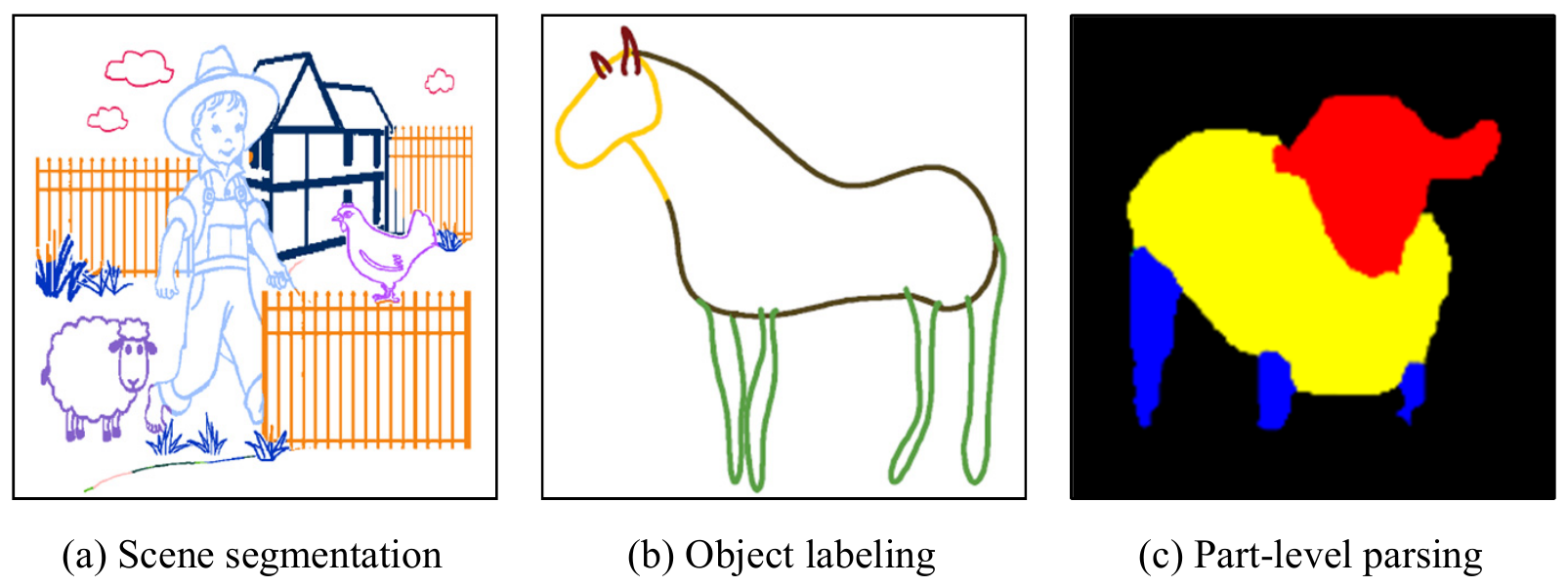}}
    \caption{Illustration of different tasks of sketch parsing. (a) shows the output of scene segmentation taken from \cite{zou2018sketchyscene}, (b) is the result of object labeling taken from \cite{schneider2016example}, (c) presents the result of our method. (a) and (b) are stroke-level labeling, while (c) is part-level parsing. Different colors refer to specific semantic classes.}
    \label{fig:sketchparsing}
\end{figure}

\textbf{Stroke-level labeling.} Most of the existing works in the field of semantic sketch parsing focus on the task of stroke-level labeling, where the goal is to make predictions inferring labels for every stroke or line segment of the freehand sketch. According to the target difference, this task can be divided into two types: scene segmentation and object labeling. The former takes the scene sketch as the input and segments all strokes of the scene into different semantic objects \cite{sun2012free} \cite{zou2018sketchyscene}. The latter labels the strokes of an individual object sketch with classes, which correspond to different semantic object parts \cite{li2018universal} \cite{huang2014data} \cite{qi2015making} \cite{wu2018sketchsegnet}. Fig. \ref{fig:sketchparsing} (a) and (b) illustrate examples for these two types of task in the stroke-level labeling. However, as the goal of this paper is different from these works, it remains unknown how to transfer such methods to solve the problem of part-level parsing for freehand sketches.

\textbf{Part-level parsing.} Unlike stroke-level labeling, the goal of part-level parsing is to predict class labels for every pixel instead of only the stroke, as shown in Fig. \ref{fig:sketchparsing} (c). From the view of output, it is similar to the part-level parsing of real images. Sarvadevabhatla et al. \cite{sarvadevabhatla2017sketchparse} firstly propose this task and collect the SketchParse dataset for the evaluation of parsing models. They present a two-level fully convolutional network and incorporate the pose prediction as an auxiliary task to provide supplementary information. To reduce the domain gap between real images and freehand sketches, they translate the real image to sketch-like form based on the edge map. However, it is just a simple expedient, which leaves much room for improvements. In this paper, we propose a homogeneous transformation method that has been experimentally proven very effective for this problem. Furthermore, we present a soft-weighted loss function and staged learning strategy to further improve the parsing performance.

\subsection{Domain Adaptation}
When the model trained on source data of a specific domain is applied to target data from another different domain, the distribution variation between two domains usually degrades the performance at the testing time \cite{patel2015visual} \cite{zhao2017continuous}. To solve this problem, domain adaption is a promising solution and has been recognized as an essential requirement. There are lots of domain adaptation methods that have proven to be successful for various fields of computer vision, such as image classification \cite{bergamo2010exploiting} \cite{saenko2010adapting}, object detection \cite{xu2018webly} \cite{rozantsev2018beyond}, and semantic image parsing \cite{van2015transfer} \cite{hong2016learning}. Saenko et al. \cite{saenko2010adapting} propose to adapt visual category models to new domains for image recognition by learning a transformation in the feature distribution. Instead of learning features that are invariant to the domain shift, Rozantsev et al. \cite{rozantsev2018beyond} state that explicitly modeling the shift between two domains should be more effective. Unlike most of these works that the modality of images actually does not change, the task of this paper faces a modality-level variation (real image vs freehand sketch), which makes it even more challenging. The challenge also exists in other sketch related fields (e.g. sketch-based image retrieval), in which the existing methods usually take the edge maps of real images as a similar data form to the freehand sketch \cite{song2017deep} \cite{yu2016sketch} \cite{sangkloy2016sketchy}. Different from these methods, we propose a homogeneous transformation method that transforms the data of two different domains into a homogeneous space to minimize the semantic gap.

\section{Methods}
\label{sec:methods}
We present a novel deep semantic sketch parsing (DeepSSP) framework for the part-level dense prediction of freehand sketches, which incorporates three new methods that solve the existing problems from different angles. In this section, we first formulate the part-level sketch parsing problem, and then show details of the proposed homogeneous transformation method, soft-weighted loss function, and staged learning strategy.

\subsection{Problem Formulation}
Part-level sketch parsing aims to decompose freehand sketches into semantic parts, which requires dense prediction of every pixel in sketches. Formally, given a freehand sketch $S$ with pixel size $h \times w$, our goal is to learn a deep parsing model $\mathcal{P}$ that can effectively segment $S$ and output an accurate semantic segmentation mask $\mathcal{O}$ with the same size as $S$. In the segmentation mask $\mathcal{O}$, $\mathcal{O}(i, j)$ means the predicted part class label for the pixel at location $(i, j)$.

As paired sketch and ground truth annotations $\left \langle S, G \right \rangle$ are not available, we utilize the paired data $\left \langle I, G \right \rangle$ from datasets of part-level real image parsing to learn the parsing model $\mathcal{P}$. Then the problem of part-level sketch parsing can be formulated as,
\begin{equation}
	\begin{split}
        \mathcal{O} = \mathcal{P}_\theta(x)
    \end{split}
    \label{eq:cross_entropy_loss}
\end{equation}
in which $\theta$ is the parameters of $\mathcal{P}$, and
\begin{equation}
	\begin{split}
        x =
        \begin{cases}
            I,& real~image\\
            S,& freehand~sketch
        \end{cases}
    \end{split}
    \label{eq:cross_entropy_loss}
\end{equation}
When transferring $\mathcal{P}$ trained on $I$ to parse $S$, minimizing the gap $\mathcal{D}(I, S)$ between two domains is one of key factors for higher performance.

\subsection{Homogeneous Transformation}
Before introducing the proposed homogeneous transformation method, we first present a glance at the problem of domain adaptation. Given data from one domain $A$ for model training, the target is to make predictions in another domain $B$. As domain $B$ has a different distribution with $A$, it makes the trained model difficult to obtain satisfactory performance. To solve this problem, methods for domain adaptation are undoubtedly necessary. In the sketch related fields, one of the most frequently-used methods to reduce the domain gap between freehand sketches and real images is to convert real images ($A$) into edge maps ($A^\ast$). The data from the new domain $A^\ast$ looks more like the freehand sketch ($B$) than the original real image, making it easier to train a better model. However, there still remains an obvious difference between the domain $A^\ast$ and $B$.

\begin{figure}[h]
    \centering
    \centerline{\includegraphics[width=0.6\linewidth]{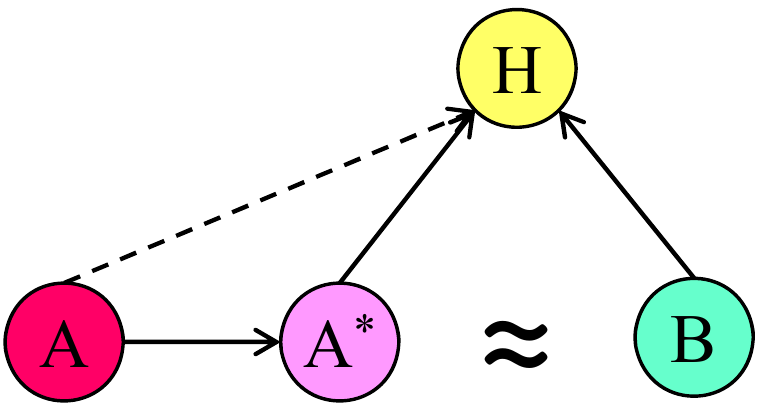}}
    \caption{Illustration of the proposed homogeneous transformation. $A$ and $B$ are two different domains. The domain $A^\ast$ that converted from domain $A$ is similar to domain $B$. $H$ is the homogeneous space, in which samples translated from domain $A^\ast$ and $B$ represent the same property.}
    \label{fig:ht}
\end{figure}

To take one step further, we make a new definition named ``homogeneous space'' ($H$), in which the data represent the same property regardless of the source domain. As shown in Fig. \ref{fig:ht}, space $H$ is transformed from domain $A^\ast$ and $B$. This process is called homogeneous transformation. It is possible to directly translate the data of domain $A$ into space $H$. However, considering that the domain $A^\ast$ is closer to the domain $B$ than the domain $A$, we choose $A^\ast$ as the source domain instead of $A$. When the training and prediction are conducted in space $H$, it is expectable to achieve higher performance for the trained model.

There are two important factors in homogeneous transformation. The first is the selection of property shared in the homogeneous space. The second is that the transformation method should minimize the variation of appearance related to the ground truth label. Otherwise, if the appearance shows a remarkable change, it may be inconsistent with the given label. In this paper, we choose the ``stroke thickness'' as the shared property and convert the strokes of all edge maps and freehand sketches into 1-pixel thickness. As we only change the thickness of strokes, the appearances of generated samples are guaranteed consistent after the transformation.

In practice, we first translate image $I$ from the domain $A^\ast$ or $B$ to a binary image $I_b$ by performing a threshold operation. In the experiments, we set the threshold $t=128$. Then we adopt a morph-based skeletonization method to extract the logical centerline $M$ of all binary strokes, which can be implemented by the `bwmorph' function of Matlab. This method removes pixels on boundaries of the binary image, without allowing it to break apart. The remaining pixels make up the centerline $I'$, which has 1-pixel thickness. Algorithm~\ref{code:method} summarizes the implementation process of this homogeneous transformation.
\begin{algorithm}[htbp]
\caption{Homogeneous transformation algorithm}
{\bf{Input}}: Image $I$ from domain $A^\ast$ or $B$\\
{\bf{Output}}: Image $I'$, in which all strokes are 1-pixel thickness
\begin{algorithmic}[1]
\STATE $I_b \leftarrow I < t$ ($t$ = 128); // get the binary image
\STATE $M \leftarrow$ bwmorph($I_b$, `skel', Inf); // get the logical centerline by a morph-based skeletonization method
\STATE $I' \leftarrow$ ones(size($I$)) * 255; // initiate $I'$ with all pixels equals to 255 (white color)
\STATE $I'(M)$ = 0; // set pixels belonging to the centerline to 0 (black color)
\STATE Return $I'$
\end{algorithmic}
\label{code:method}
\end{algorithm}

After this operation, all images from the source and target domains are transformed into the homogeneous space, in which all strokes of samples share the same property. Finally, the training and test images are replaced with their corresponding samples in the homogeneous space. As the proposed homogeneous transformation is not restricted to the task of part-level semantic sketch parsing, it can be taken as a general method for sketch related applications, such as sketch-based image retrieval.

\subsection{Soft-Weighted Loss}
The soft-weighted loss is designed to address the part-level sketch parsing scenario, in which there are ambiguous label boundary and class imbalance between different semantic parts during training. Before introducing the soft-weighted loss, we first start from the definition of standard cross entropy (CE) loss for each pixel,
\begin{equation}
	\begin{split}
        l(x, i) &= -log \left( e^{x_i} \bigg / \sum \limits_{j=0}^{C-1} e^{x_j} \right)\\
                &= -x_i + log \sum \limits_{j=0}^{C-1} e^{x_j}
    \end{split}
    \label{eq:cross_entropy_loss}
\end{equation}
where $x$ is the input that contains predicted scores for each class, $i$ is the ground truth class label, $C$ refers to the number of classes. The final CE loss for each prediction of the part-level semantic sketch parsing is computed by,
\begin{equation}
	\begin{split}
        \mathcal{L} = \frac {\sum _p l(x_p, i_p)} {w \times h}
    \end{split}
    \label{eq:final_loss}
\end{equation}
which averages losses at all positions $p$ of the prediction with the resolution of $w \times h$.

We propose a soft-weighted loss to address the problems of ambiguous label boundary and class imbalance, which reshapes the standard CE loss to the following formulation,
\begin{equation}
	\begin{split}
        l_s(x, i) &= \alpha_i \sum \limits_{j=0}^{C-1} \lambda_j l(x, j)
    \end{split}
    \label{eq:soft_weighted_loss}
\end{equation}
where $\lambda$ is used to handle the situation of ambiguous label boundary, $\alpha$ is a weighting parameter that re-weights the losses of different classes. By substituting Eq. (\ref{eq:cross_entropy_loss}) into Eq. (\ref{eq:soft_weighted_loss}), the soft-weighted loss can be written as,
\begin{equation}
	\begin{split}
        l_s(x, i) &= \alpha_i \sum \limits_{j=0}^{C-1} \lambda_j \left( -x_j + log \sum \limits_{k=0}^{C-1} e^{x_k} \right) \\
             &= \alpha_i \left( -\sum \limits_{j=0}^{C-1} \lambda_j x_j + \sum \limits_{j=0}^{C-1} \lambda_j log \sum \limits_{k=0}^{C-1} e^{x_k} \right)\\
             &= \alpha_i \left( -\sum \limits_{j=0}^{C-1} \lambda_j x_j + log \sum \limits_{k=0}^{C-1} e^{x_k} \right)
    \end{split}
    \label{eq:soft_weighted_cross_entropy_loss}
\end{equation}
Next, we present the details of these two parameters ($\lambda$, $\alpha$) and show their specific effects on the task of part-level semantic sketch parsing.

As a high abstraction of objects or scenes, the freehand sketch lacks lots of cues (e.g., texture and color) when compared to the real image, which frequently makes the label boundary of adjacent parts ambiguous. For instance, the labels distributed over the boundary of the part class ``head'' and ``torso'' are not completely certain. The class ``head'' and ``torso'' can be seen as the soft labels for these pixels. It should be more acceptable to assign the soft label ``torso'' to the pixel labeled with ``head'' on the boundary than other labels like the ``tail'' and ``leg''. Therefore, we introduce the soft parameter $\lambda$ to give some tolerance to predictions that output the soft labels to boundary pixels instead of the ground truth classes.

\begin{figure}[h]
    \centering
    \centerline{\includegraphics[width=0.8\linewidth]{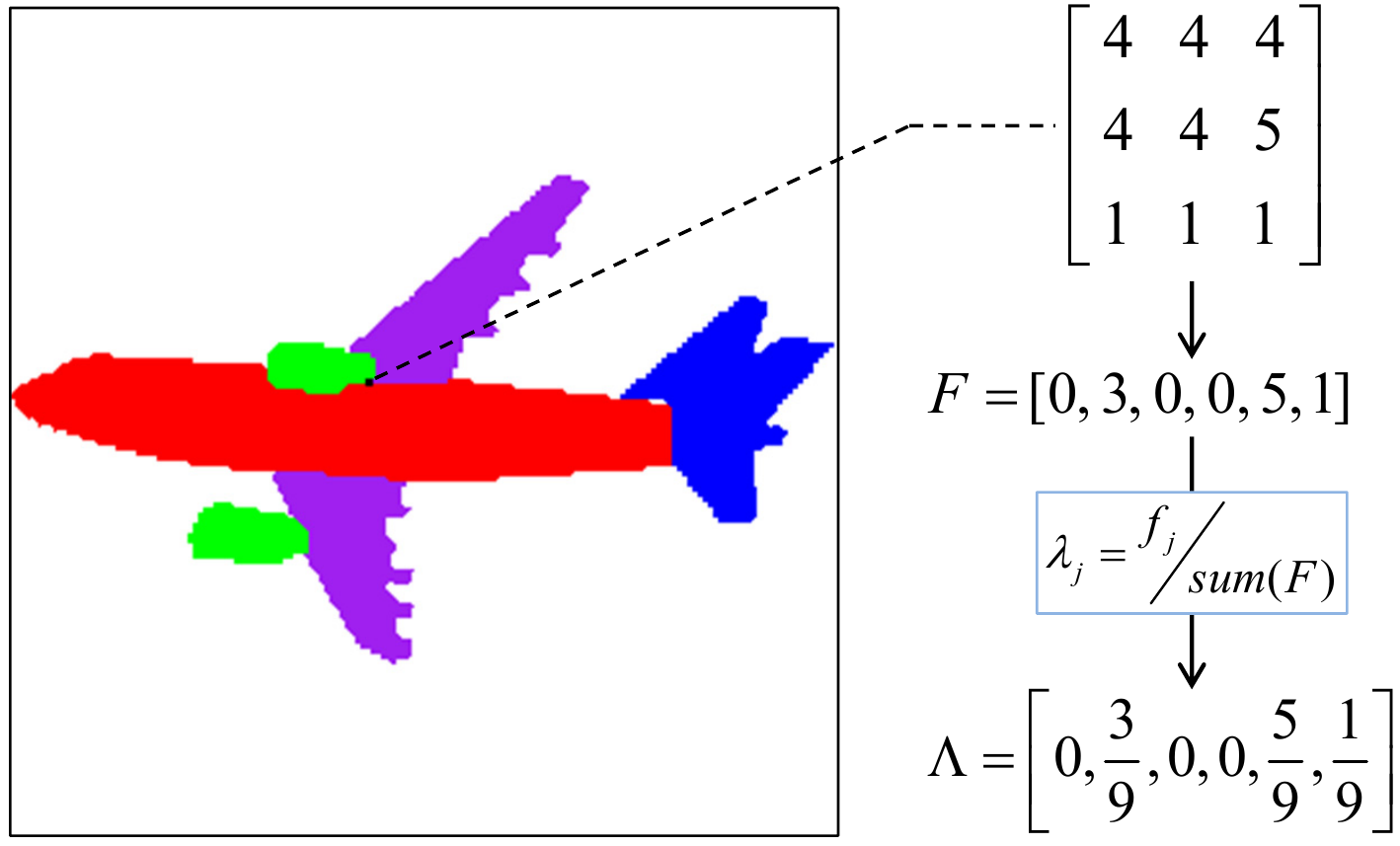}}
    \caption{Illustration of the computation of soft parameters $\lambda$ for the boundary pixel with label ``4''. The top matrix shows the class labels of adjacent pixels, $F$ counts the number of each class, $\Lambda = \{\lambda_0, \lambda_1, \ldots, \lambda_{C-1}\}$ consists of the soft parameters of all classes.}
    \label{fig:soft}
\end{figure}

The soft parameter $\lambda_j$ for class $j$ is computed by,
\begin{equation}
	\begin{split}
        \lambda_j = f_j \bigg / \sum \limits_{k=0}^{C-1} f_k
    \end{split}
    \label{eq:soft_weight}
\end{equation}
in which $f_j$ counts the number of pixels belonging to class $j$, $\lambda_j$ is equivalent to the percentage of class $j$ among these adjacent pixels. For a better understanding, we present an illustration of the computation of soft parameters for the boundary pixel, as shown in Fig. \ref{fig:soft}. In practice, we only take foreground classes into consideration and set $f_0=0$, where the background class is indexed with $0$. For pixels belonging to class $i$ that are not adjacent to other parts, $\lambda_i=1$, while for other cases $\lambda_j=0$ ($j \neq i$), which makes the soft-weighted loss evolving into,
\begin{equation}
	\begin{split}
        l_s(x, i) = \alpha_i \left( -x_i + log \sum \limits_{k=0}^{C-1} e^{x_k} \right)
    \end{split}
    \label{eq:soft_weighted_cross_entropy_loss_nomal}
\end{equation}
We can see that the soft-weighted loss focuses on adjusting the loss for boundary pixels and preserves the loss for pixels with a clear label. It makes the parsing model more concentrated on reducing losses with clear errors while avoiding the disturbance brought by the ambiguous label boundary.

Class imbalance is a common problem in the field of computer vision \cite{lin2017focal} \cite{rota2017loss}. For the task of semantic sketch parsing, there is a great difference in the distribution between different part classes. As a consequence, class with lots of pixels dominates the training loss, which brings negative impact to the model training. To alleviate this issue, we apply the weighting parameter $\alpha$ to re-weight the losses from different classes. For pixels from the part class $i$, the weighting parameter $\alpha_i$ is defined as,
\begin{equation}
	\begin{split}
        \alpha_i = \frac {M} {\varphi_i}
    \end{split}
    \label{eq:weight}
\end{equation}
where $M$ is the median of ${\varphi_0, \dots, \varphi_{C-1}}$, and $\varphi_i$ is computed as follows,
\begin{equation}
	\begin{split}
        \varphi_i = \frac {t_i} {n_i}
    \end{split}
    \label{eq:average_pixel}
\end{equation}
where $t_i$ is the total number of pixels belonging to class $i$, and these pixels are present in $n_i$ images. The $\varphi$ can be seen as the average number of pixels on the training set for each class. Eq. (\ref{eq:weight}) guarantees that the class with few pixels has a higher weight than classes with more pixels. Finally, the soft-weighted cross entropy loss is formulated as,
\begin{equation}
	\begin{split}
        \mathcal{L}_s = \frac {\sum _p l_s(x_p, i_p)} {\sum _p \alpha_{i_p}}
    \end{split}
    \label{eq:loss}
\end{equation}
in which $\alpha_{i_p}$ means the weighting parameter of position $p$ whose ground truth class label is $i$.

\begin{figure}[t]
    \centering
    \subfigure[Stage 1]{
        \begin{minipage}[t]{0.475\linewidth}
        \centering
        \includegraphics[width=1.0\linewidth]{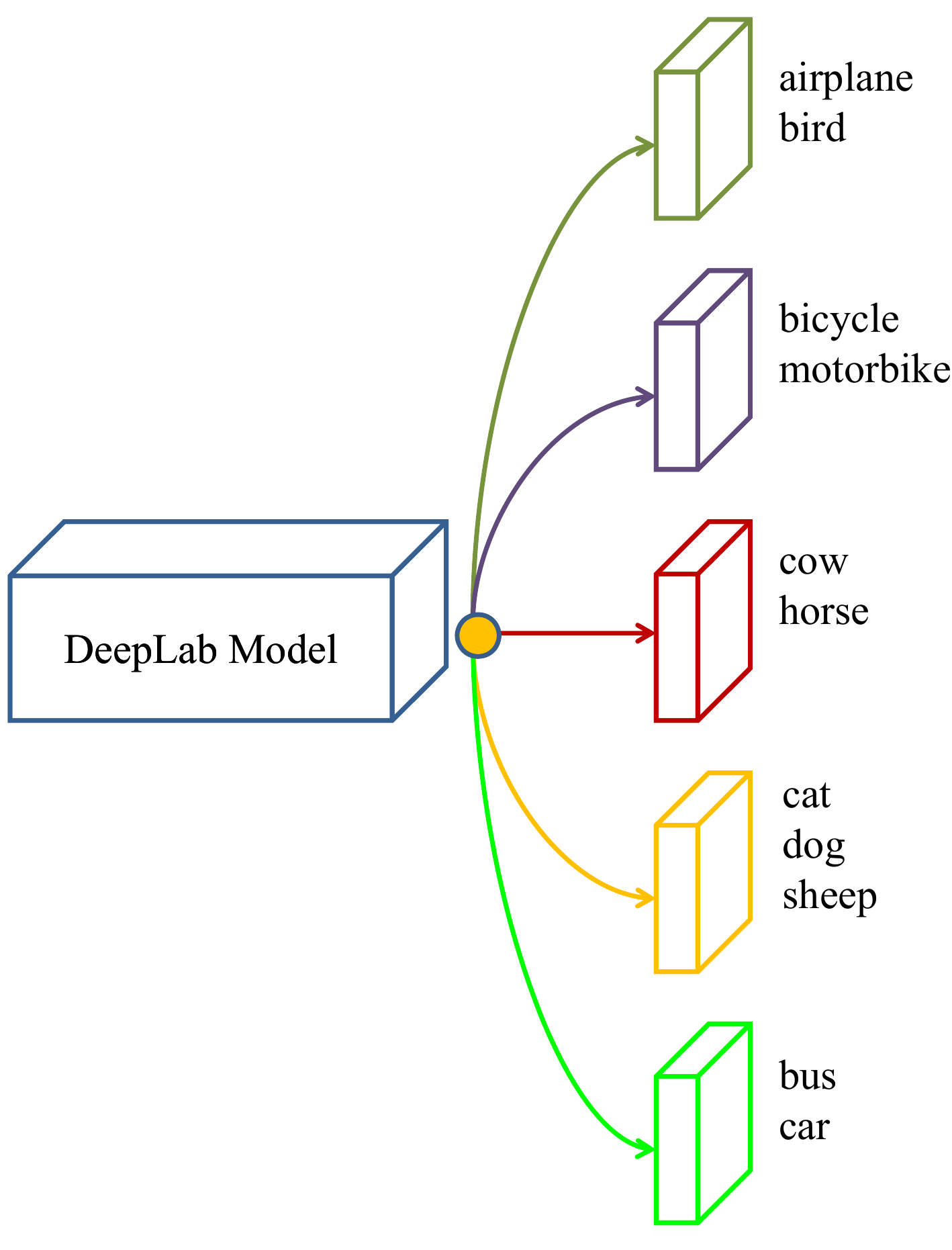}
        \end{minipage}
    }
    \subfigure[Stage 2]{
        \begin{minipage}[t]{0.455\linewidth}
        \centering
        \includegraphics[width=1.0\linewidth]{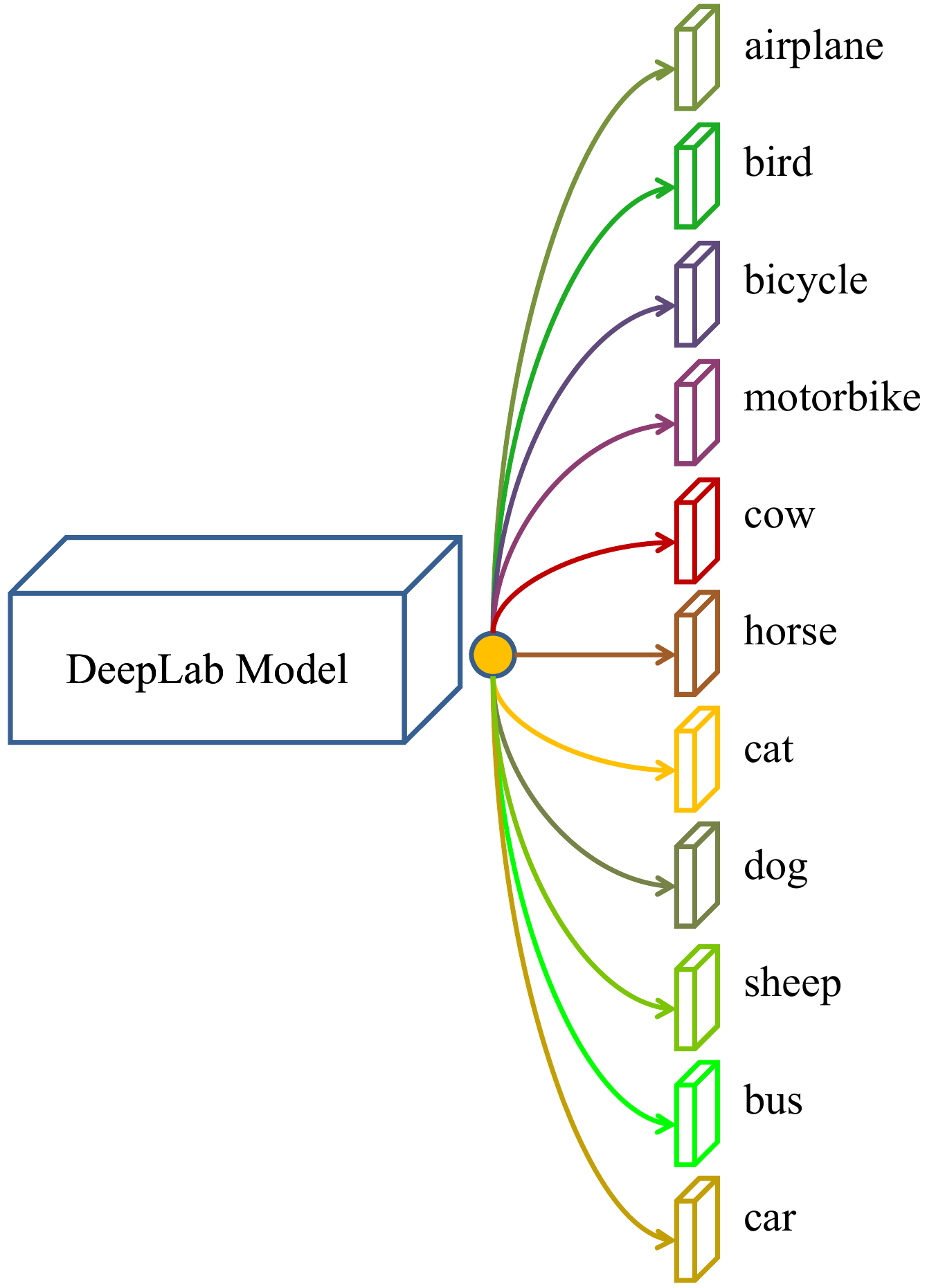}
        \end{minipage}
    }
    \caption{Our staged learning strategy for training the sketch parsing model. (a) Stage 1: the half-shared network \cite{sarvadevabhatla2017sketchparse} with 5 super branches is used to learn parameters in shared layers of the DeepLab model, in which sub-categories like ``cow'' and ``horse'' under the same super category ``large animals'' share a super branch. (b) Stage 2: we freeze these shared layers and train unshared layers of each specific category under a full branch architecture. The shared layers of the DeepLab model are shown as the left blue box at each stage, and the branches for super or specific categories are presented in right boxes with different colors.}
    \label{fig:staged_learning}
\end{figure}

\subsection{Staged Learning}
\label{sec:staged_learning}
Given the category of freehand sketches, an intuitive way for semantic sketch parsing is to train the network independently for each category. It is straightforward for training but neglects the information shared across different categories. Their performance is greatly limited when only a small number of training samples are available. This problem can be alleviated via a half-shared super branch architecture \cite{sarvadevabhatla2017sketchparse}, which segments the semantic parsing model into two parts, as shown in Fig. \ref{fig:staged_learning} (a). The front part consists of several shared layers of the DeepLab model, while the rest layers are heterogeneous for 5 super branches. Each super branch is designed to achieve the goal of parsing sketches from the related super category, in which sub-categories such as ``cow'' and ``horse'' have similar semantic part classes. The super categories and their corresponding sub-categories defined in \cite{sarvadevabhatla2017sketchparse} are presented in Table \ref{tab:categories}. However, this super branch network does not consider the difference between sub-categories under the same super category.

\begin{table}[htbp]
\centering
\caption{Super categories and their corresponding sub-categories.}
\begin{tabular}{ccccccc}
\toprule
\multicolumn{2}{c}{\textbf{large animals}} & \multicolumn{2}{c}{\textbf{4-wheelers}} & \multicolumn{3}{c}{\textbf{small animals}}\\
{cow} & {horse} & {bus} & {car} & {cat} & {dog} & {sheep}\\
\midrule
\multicolumn{2}{c}{\textbf{2-wheelers}} & \multicolumn{2}{c}{\textbf{flying things}}\\
{bicycle} & {motorbike} & {airplane} & {bird}\\
\bottomrule
\end{tabular}
\label{tab:categories}
\end{table}

In consideration of the information sharing and specific characteristic for each sketch category, we propose a staged learning strategy to further improve the parsing performance of the trained model. As shown in Fig. \ref{fig:staged_learning}, the strategy consists of two training stages that are independent of backbone networks. At stage 1, we use training samples from all sketch categories to learn the parameters of shared layers $\mathcal{S}$ under the half-shared super branch architecture. In each iteration, the data flow forwards from shared layers to their corresponding branch layers. At stage 2, we freeze all shared layers and replace each super branch with several sub-branches, as shown in Fig. \ref{fig:staged_learning} (b). In this stage, we only need to fine-tune layers of the corresponding branch for each sketch category. Experimental results demonstrate the superior performance of our strategy compared to the complete independent training and super branch architecture.

\section{Experiments}
\label{sec:experiments}
In this section, we first introduce datasets used for the training and evaluation. Then, we give details of the experimental implementation and propose a novel erasing-based method for data augmentation. To provide a more comprehensive understanding of the proposed method, we evaluate the contributions of each component and present an ablation study via extensive experiments. Furthermore, extra experiments are conducted on the task of fine-grained sketch-based image retrieval to demonstrate the practical value of the homogeneous transformation method. Finally, we present a comparison against other existing methods and make discussions of some qualitative results.

\subsection{Datasets and Evaluation Metric}
\begin{figure}[htbp]
    \centering
    \centerline{\includegraphics[width=1.0\linewidth]{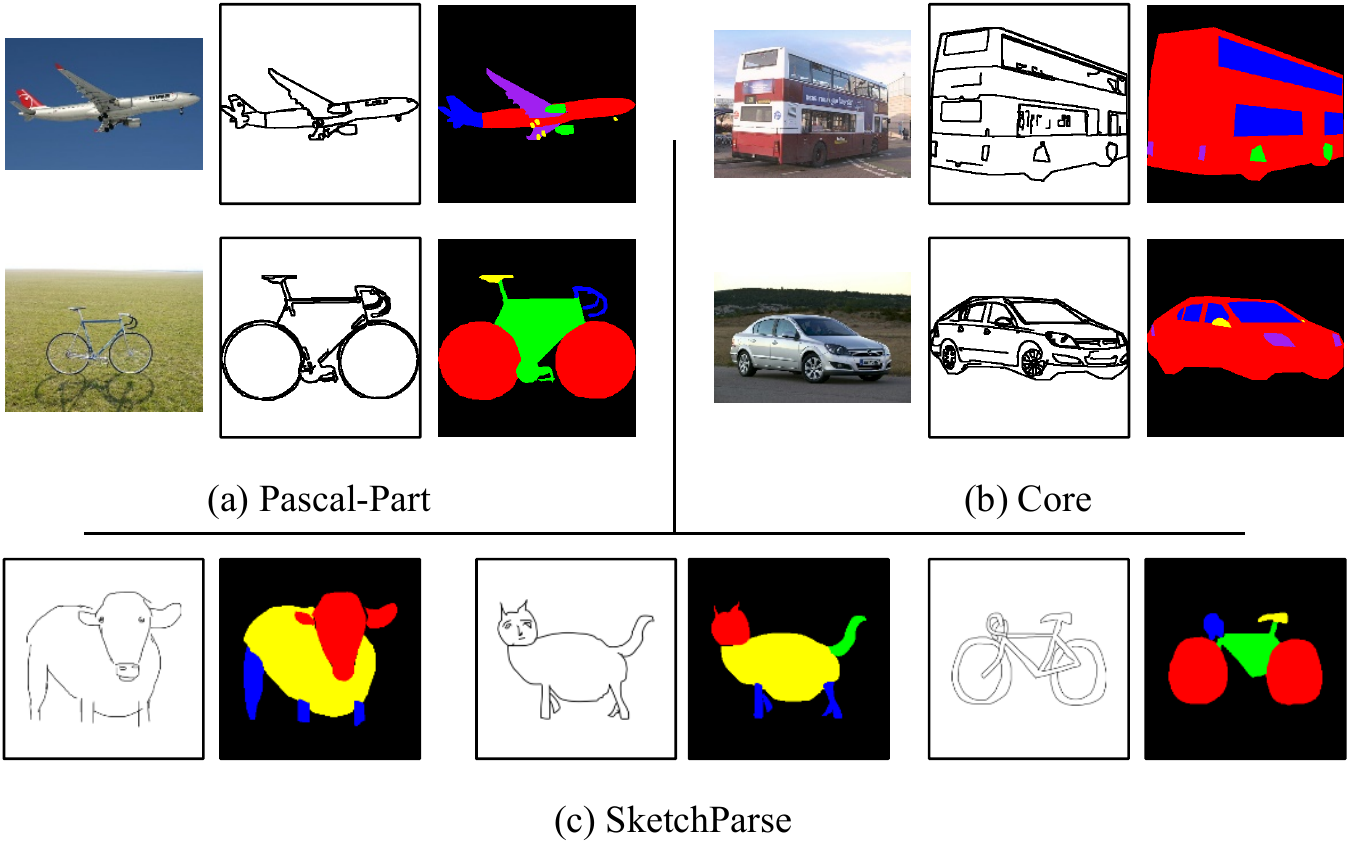}}
    \caption{Examples from the Pascal-Part, Core, and SketchParse datasets.}
    \label{fig:datasets}
\end{figure}

Following the work of Sarvadevabhatla et al. \cite{sarvadevabhatla2017sketchparse}, we use the data from real image datasets for model training and evaluate the performance of trained models on the SketchParse dataset. Specifically, the training set consists of 1532 paired real images and corresponding part-level annotations, which distribute across 11 categories (i.e., airplane, bicycle, bird, bus, car, cat, cow, dog, horse, motorbike, and sheep). These images and annotations are selected from two public datasets, i.e., Pascal-Part \cite{chen2014detect} and Core \cite{farhadi2010attribute}. Instead of edge maps, we apply the proposed homogeneous transformation on sketchified images provided by \cite{sarvadevabhatla2017sketchparse}, which are jointly generated by the Canny edge detector and ground truth part-annotations. Examples of real images, sketchified images, and their part-level annotations are illustrated in Fig. \ref{fig:datasets} (a) and (b).

The evaluation is conducted on the SketchParse dataset \cite{sarvadevabhatla2017sketchparse}, which takes 48 freehand sketches for each category respectively from the Sketchy \cite{sangkloy2016sketchy} and TU-Berlin \cite{eitz2012humans} datasets. As the category ``bus'' only exists in the TU-Berlin dataset, there are totally 1008 ($48 \times 2 \times 10 + 48$) freehand sketches in the SketchParse dataset. All sketches are labeled with part-level dense annotations, as shown in Fig. \ref{fig:datasets} (c).

The same with \cite{sarvadevabhatla2017sketchparse}, we take the average IOU score as the quantitative metric to evaluate the parsing performance of trained models. For each class $i$, the IOU score is computed by,
\begin{equation}
	\begin{split}
        \mathcal{P}_i = \frac {n_{ii}} {\sum\nolimits_j n_{ij} + \sum\nolimits_j n_{ji} - n_{ii}}
    \end{split}
    \label{eq:score_p}
\end{equation}
where $n_{ij}$ is the number of pixels predicted to class $j$ among the pixels belonging to class $i$. Then, the IOU score on a single sketch is computed by,
\begin{equation}
	\begin{split}
        \mathcal{I} = \frac {\sum\nolimits_i \mathcal{P}_i} {n_p}
    \end{split}
    \label{eq:score_i}
\end{equation}
where $n_p$ is the number of unique part labels in the sketch. By averaging the IOU scores on all test sketches, we can get the final average IOU score as follows,
\begin{equation}
	\begin{split}
        \mathcal{A} = \frac {\sum\nolimits \mathcal{I}} {n_s}
    \end{split}
    \label{eq:score_i}
\end{equation}
in which $n_s$ is the number of test sketches.

\subsection{Implementation Details}
We take DeepLab v2 \cite{chen2018deeplab} as the backbone network, which is a widely used architecture for semantic parsing. In the experiments, the DeepLab model is derived from a multi-scale version of ResNet-101 \cite{he2016deep}. Similar to the work of \cite{sarvadevabhatla2017sketchparse}, we split the deep model into two parts at the position of ``res5b''. As shown in Fig. \ref{fig:staged_learning}, the front part is used as shared layers across categories and the rest layers are copied into different branches. The initial learning rate is set to $5 \times 10^{-4}$ except for the final convolutional layers. The learning rate of the final convolutional layers in each branch is set to $5 \times 10^{-3}$. The learning rate is changed under the polynomial decay policy. Limited by the memory of GPU, the mini-batch size is set to 1. We adopt the stochastic gradient descent (SGD) with a momentum of 0.9 as the optimizer, in which the weight decay is set to 0 by default. Furthermore, we apply 20000 iterations to learn the parameters of shared layers at stage 1 and 2000 iterations to fine-tune the rest layers for each branch at stage 2. All experiments are conducted on a single NVIDIA GeForce GTX 1080Ti GPU with 11GB memory.

Data augmentation is an important strategy to improve the performance of deep neural networks \cite{zhong2020random} \cite{wang2017fast}. Same as \cite{sarvadevabhatla2017sketchparse}, we perform different degrees of rotations (0, $\pm$10, $\pm$20, $\pm$30) and mirroring on the original image, which finally outputs 14 augmented images for each sketch. Furthermore, we apply an erasing-based sketch augmentation method to generate two times of training data. For each training image with the resolution of $321 \times 321$ pixels, the erasing augmentation method randomly erases a region with a size of $31 \times 31$ pixels. The generated images share the ground truth annotation with their source images.

\subsection{Ablation Study}
\begin{figure}[htbp]
    \centering
    \subfigure[Different initial learning rates]{
        \begin{minipage}[t]{0.465\linewidth}
        \centering
        \includegraphics[width=1.0\linewidth]{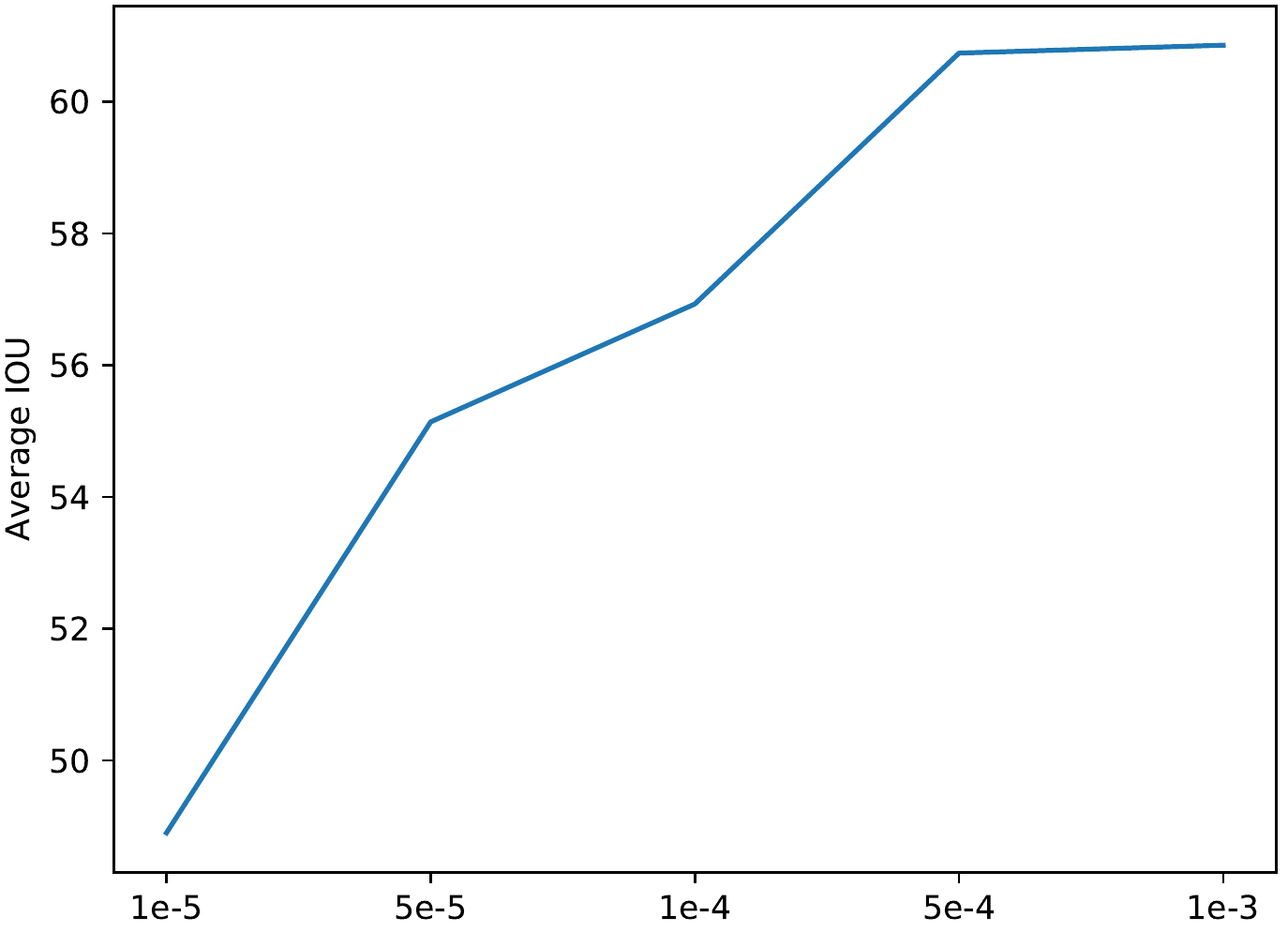}
        \end{minipage}
    }
    \subfigure[Different optimizers]{
        \begin{minipage}[t]{0.465\linewidth}
        \centering
        \includegraphics[width=1.0\linewidth]{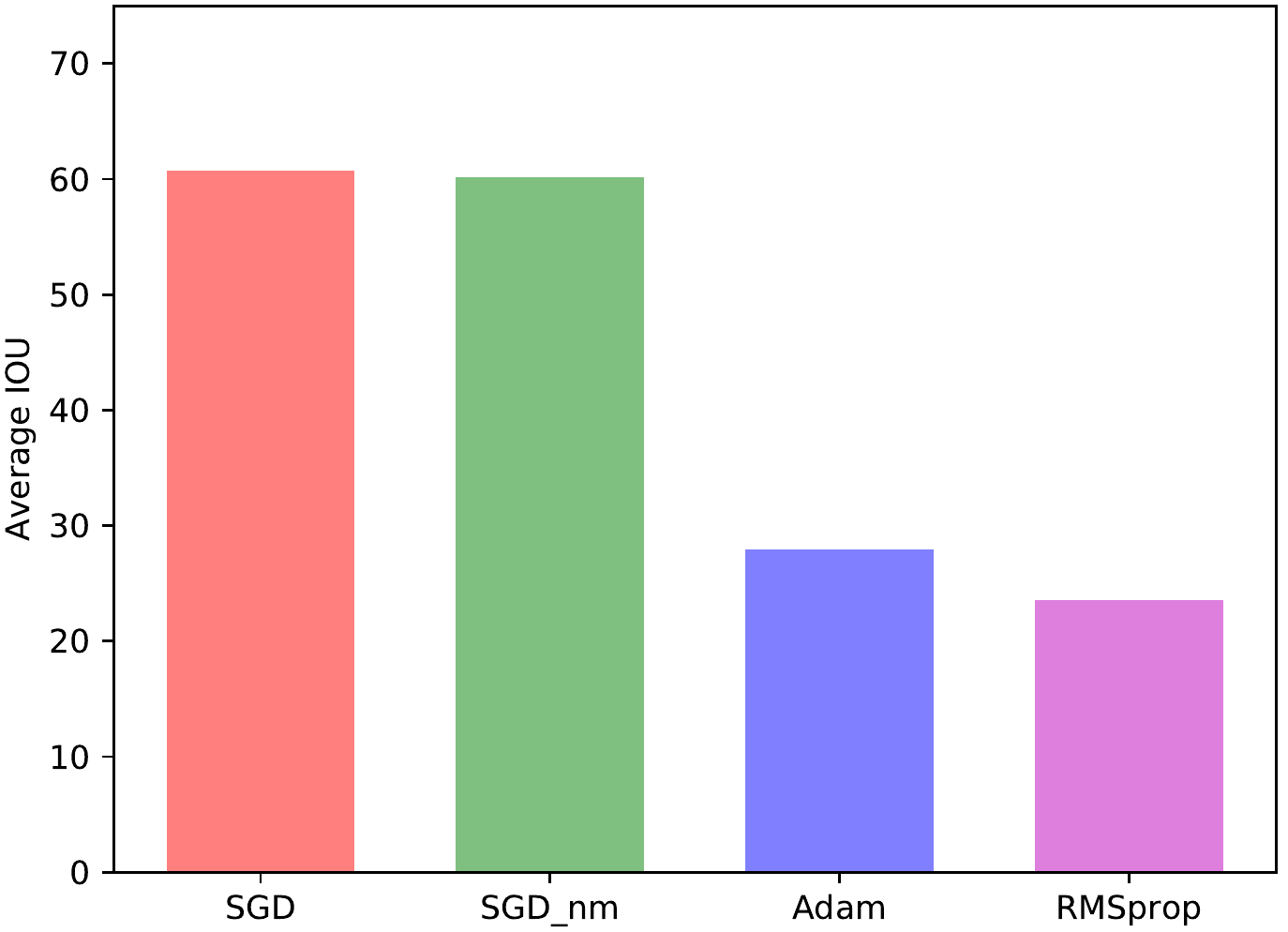}
        \end{minipage}
    }
    \caption{Comparison of different initial learning rates and optimizers.}
    \label{fig:different_lr_opt}
\end{figure}

Before evaluating each component of our DeepSSP framework, we first analyze the impact of using different initial learning rates $lr$ and optimizers on performance. We conduct experiments with the same baseline model, which is trained with the rotation and mirroring augmentation, the standard cross entropy loss, and the super branch architecture. Fig. \ref{fig:different_lr_opt} (a) presents the comparison results of different initial learning rates. It can be seen that when setting $lr$ with a small value like $1e-5$, the average IOU score is very low. With the increase of $lr$, from $1e-5$ to $5e-4$, the performance grows rapidly. Although the performance of $lr=1e-3$ is slightly better than $lr=5e-4$, we find that $lr \geq 1e-3$ makes the training process very unstable and hard to converge. Therefore, we set the initial learning rate $lr=5e-4$ (i.e., $5 \times 10^{-4}$), which outputs a good and stable performance.

\begin{table*}[t]
\centering
\caption{Comparative results of using different training or test data on the SketchParse dataset.}
\begin{tabular}{lcccccccccccc}
\toprule
{} & {~~cow~~} & {~~horse~~} & {~~cat~~} & {~~dog~~} & {~~sheep~~} & {~~bus~~} & {~~car~~} & {~bicycle~} & {~motorbike~} & {~airplane~} & {~~bird~~} & {~~average~~}\\
\midrule
{Original} & 64.39 & 66.23 & \textbf{65.01} & 66.05 & 64.63 & 63.55 & 57.97 & 50.60 & 50.20 & 53.13 & 43.25 & 58.37\\
{HT$^-$} & 61.64 & 64.45 & 61.36 & 61.97 & 63.16 & 61.76 & 60.84 & 53.96 & 50.07 & 54.28 & 44.16 & 57.76\\
{HT} & \textbf{64.70} & \textbf{66.72} & 63.88 & \textbf{67.16} & \textbf{66.63} & \textbf{65.40} & \textbf{65.94} & \textbf{57.87} & \textbf{50.88} & \textbf{55.71} & \textbf{45.70} & \textbf{60.72}\\
\bottomrule
\end{tabular}
\begin{tablenotes}
\item $^{\ast}$ ``Original'' means using the original training data. ``HT$^-$'' trains the model with the training data after the homogeneous transformation but predicts on the test data without HT. ``HT'' refers that both the training and evaluation are conducted on the data with HT. For a pure evaluation of these methods, all models are trained without any data augmentation.
\end{tablenotes}
\label{tab:ht_comp}
\end{table*}

We also compare the performance of four different optimizers: SGD (with momentum), SGD\_nm (without momentum), Adam, and RMSprop. For Adam, the Beta and weight decay are set to (0.9, 0.99) and 0 respectively. For RMSprop, the Alpha is set to 0.9, while the weight decay and momentum are set to 0 by default. As shown in Fig. \ref{fig:different_lr_opt} (b), the average IOU score of SGD is better than SGD\_nm (60.74\% vs 60.15\%), which proves the effectiveness of momentum when optimizing network with SGD. From the figure, it can be seen that the SGD is significantly better than the other two optimizers (Adam and RMSprop). Therefore, we choose SGD as the optimizer in all experiments.

In the experiments, we apply the homogeneous transformation (HT) method for the training and test dataset. Then both the model training and evaluation are conducted on the transformed data. We select the super branch architecture as the base network and present the results in Table \ref{tab:ht_comp}. Compared to the training on the original dataset, the model with our HT method achieves better performance among 10 of 11 sketch categories and gets 2.35\% higher IOU score on average. The results demonstrate the effectiveness of the proposed HT method on the task of part-level semantic sketch parsing. Furthermore, we show the results of only performing the HT method on the training data but do nothing on the test set, shown as ``HT$^{-}$'' in Table \ref{tab:ht_comp}. In this case, there still is a big gap between the training and test datasets, which gets the worst performance. Therefore, it is important to perform the HT method on both domains simultaneously.

We compare the performance of different augmentation methods and combinations in Table \ref{tab:ht_improve}. It can be seen that the rotation-based augmentation shows the best performance among these three methods when used in isolation. By combining them together, the average IOU score achieves 2.92\% gain compared to the vanilla version (from 58.37\% to 61.29\%). Therefore, we adopt this combination of augmentation methods for model training. We also present the performance improvements brought by the homogeneous transformation in combination with these augmentation combinations. As shown in Table \ref{tab:ht_improve}, we can see that the HT method invariably outperforms the training with the original dataset, which demonstrates the stability and effectiveness of the proposed HT method.

\begin{table}[htbp]
\centering
\caption{Comparative results of different data augmentation methods and performance improvements brought by the homogeneous transformation (HT).}
\begin{tabular}{lccc}
\toprule
{} & {~~~Original~~} & {~~~HT~~} & {~~Improvement~~}\\
\midrule
{vanilla} & 58.37 & \textbf{60.72} & +2.35\\
{mirroring} & 59.04 & \textbf{61.34} & +2.30\\
{rotation} & 60.46 & \textbf{61.72} & +1.26\\
{erasing} & 59.74 & \textbf{60.86} & +1.12\\
{mirroring+rotation} & 60.74 & \textbf{61.69} & +0.95\\
{erasing+mirroring} & 60.02 & \textbf{61.83} & +1.81\\
{erasing+rotation} & 60.94 & \textbf{61.98} & +1.04\\
{erasing+mirroring+rotation} & 61.29 & \textbf{62.71} & +1.42\\
\bottomrule
\end{tabular}
\label{tab:ht_improve}
\end{table}

\begin{table*}[t]
\centering
\caption{Comparison of different loss functions on the SketchParse dataset.}
\begin{tabular}{lcccccccccccc}
\toprule
{} & {~~cow~~} & {~~horse~~} & {~~cat~~} & {~~dog~~} & {~~sheep~~} & {~~bus~~} & {~~car~~} & {bicycle} & {motorbike} & {airplane} & {~~bird~~} & {~~average~~}\\
\midrule
{Base+$\mathcal{L}$} & 69.37 & 71.18 & 66.91 & 67.83 & 69.12 & 65.73 & 66.33 & 58.48 & 50.79 & 57.05 & 48.80 & 62.71\\
{Base+$\mathcal{L}_{w}$} & 69.45 & 70.53 & 69.24 & 71.53 & 69.70 & 66.99 & 70.11 & 61.08 & 53.15 & 57.67 & 50.41 & 64.39\\
{Base+$\mathcal{L}_{s}$} & 69.76 & 71.30 & \textbf{70.19} & 71.30 & 71.22 & 66.98 & 69.98 & 61.60 & \textbf{56.18} & 59.14 & \textbf{52.11} & 65.33\\
{Base+$\mathcal{L}_{s}^\ast$~~~~~~~} & \textbf{70.16} & \textbf{72.12} & 69.58 & \textbf{72.03} & \textbf{71.74} & \textbf{67.18} & \textbf{70.15} & \textbf{62.35} & 54.48 & \textbf{60.45} & 52.07 & \textbf{65.57}\\
\bottomrule
\end{tabular}
\begin{tablenotes}
\item $^{\ast}$ ``Base'': training with the augmentation methods mentioned above, ``$\mathcal{L}$'': the standard cross entropy (CE) loss, ``$\mathcal{L}_w$'': the weighted CE loss, ``$\mathcal{L}_{s}$: the soft-weighted CE loss, ``$\mathcal{L}_{s}^\ast$'': the soft-weighted CE loss with a higher weight for the background class.
\end{tablenotes}
\label{tab:loss_comp}
\end{table*}

\begin{table*}[t]
\centering
\caption{Comparison of different architectures on the SketchParse dataset.}
\begin{tabular}{lcccccccccccc}
\toprule
{} & {~~cow~~} & {~~horse~~} & {~~cat~~} & {~~dog~~} & {~~sheep~~} & {~~bus~~} & {~~car~~} & {bicycle} & {motorbike} & {airplane} & {~~bird~~} & {~~average~~}\\
\midrule
{\textit{\textbf{w/o improvements}}}\\
{Independent} & 62.86 & 65.32 & 62.73 & 62.92 & 64.79 & 64.75 & 61.91 & 57.83 & 49.70 & 52.49 & \textbf{49.35} & 59.25\\
{Full Branch} & 65.03 & 65.71 & 63.31 & 65.17 & 65.60 & 65.06 & 64.34 & 58.75 & 50.22 & \textbf{54.53} & 45.24 & 60.02\\
{Super Branch} & 66.01 & 67.77 & 66.37 & 67.41 & 67.37 & 65.80 & 63.15 & \textbf{59.15} & 50.43 & 52.95 & 44.57 & 60.74\\
{Staged Learning} & \textbf{66.92} & \textbf{69.74} & \textbf{67.59} & \textbf{69.33} & \textbf{67.73} & \textbf{66.17} & \textbf{63.80} & 59.03 & \textbf{50.88} & 54.09 & 48.08 & \textbf{61.90}\\
\midrule
{\textit{\textbf{w/ improvements}}}\\
{Independent} & 66.54 & 70.02 & 66.63 & 69.94 & 68.58 & 65.95 & 67.01 & 63.09 & 51.89 & 55.97 & 52.20 & 63.30\\
{Full Branch} & 69.73 & 70.73 & 68.54 & 71.33 & 69.44 & \textbf{68.43} & 69.71 & 62.80 & 51.20 & 59.41 & 51.83 & 64.64\\
{Super Branch} & 70.16 & 72.12 & 69.58 & 72.03 & 71.74 & 67.18 & 70.15 & 62.35 & 54.48 & \textbf{60.45} & 52.07 & 65.57\\
{Staged Learning} & \textbf{70.42} & \textbf{72.81} & \textbf{69.94} & \textbf{72.57} & \textbf{72.02} & 67.88 & \textbf{70.88} & \textbf{63.30} & \textbf{55.69} & 59.96 & \textbf{54.38} & \textbf{66.25}\\
\bottomrule
\end{tabular}
\begin{tablenotes}
\item $^{\ast}$ ``Independent'': training independently for each category, ``Full Branch'': the half-shared network with one branch for each category, ``Super Branch'': the super branch architecture, ``Staged Learning'': our staged learning strategy. ``improvements'': training with the proposed erasing augmentation, homogeneous transformation, and soft-weighted loss.
\end{tablenotes}
\label{tab:staged_comp}
\end{table*}

\begin{table*}[t]
\centering
\caption{The top K accuracies (acc.@$K=1,\dots,10$) after implementing the proposed HT method on DSSA \cite{song2017deep}.}
\begin{tabular}{clcccccccccc}
\toprule
{} & {} & {~~~~1~~~~} & {~~~~2~~~~} & {~~~~3~~~~} & {~~~~4~~~~} & {~~~~5~~~~} & {~~~~6~~~~} & {~~~~7~~~~} & {~~~~8~~~~} & {~~~~9~~~~} & {~~~~10~~~~}\\
\midrule
\multirow{2}{*}{~~~Shoe~~~} & {~~~DSSA \cite{song2017deep}~~~} & 58.26 & 68.70 & 74.78 & 79.13 & 82.61 & 85.22 & 85.22 & 88.70 & 90.43 & 92.17\\
{} & {~~~DSSA+HT~~~} & \textbf{66.09} & \textbf{73.91} & \textbf{79.13} & \textbf{85.22} & \textbf{88.70} & \textbf{91.30} & \textbf{92.17} & \textbf{93.04} & \textbf{93.04} & \textbf{93.04}\\
\midrule
\multirow{2}{*}{~~~Chair~~~} & {~~~DSSA \cite{song2017deep}~~~} & 79.38 & 85.57 & 86.60 & 89.69 & 92.78 & 93.81 & \textbf{95.88}  & \textbf{95.88} & \textbf{95.88} & \textbf{95.88}\\
{} & {~~~DSSA+HT~~~} & \textbf{85.57} & \textbf{90.72} & \textbf{91.75} & \textbf{93.81} & \textbf{94.85} & \textbf{94.85} & \textbf{95.88} & \textbf{95.88} & \textbf{95.88} & \textbf{95.88}\\
\midrule
\multirow{2}{*}{~~~Handbag~~~} & {~~~DSSA \cite{song2017deep}~~~} & 48.21 & 58.33 & 66.07 & 69.05 & 73.21 & 76.79 & \textbf{79.17} & \textbf{80.95} & \textbf{82.74} & \textbf{83.33}\\
{} & {~~~DSSA+HT~~~} & \textbf{50.60} & \textbf{63.10} & \textbf{70.24} & \textbf{73.21} & \textbf{75.60} & \textbf{77.38} & 78.57 & 79.76 & 81.55  & \textbf{83.33}\\
\bottomrule
\end{tabular}
\label{tab:topkdssa}
\end{table*}

Table \ref{tab:loss_comp} shows comparative results of different loss functions on the SketchParse dataset. All models are trained with the combination of three augmentation methods mentioned above, noted as ``Base'' in the table. Compared to the standard cross entropy (CE) loss $\mathcal{L}$ and the weighted version $\mathcal{L}_w$, the model trained with the proposed soft-weighted CE loss $\mathcal{L}_s$ achieves better performance. The results show the superiority of the soft-weighted CE loss in the task of part-level semantic sketch parsing. As the pixels belonging to the background class are mostly separated from other classes by the sketch boundary, we set the weighting parameter $\alpha_0^\ast=2\alpha_0$ to make the network more sensitive to the boundary between foreground classes and the background class. As shown in Table \ref{tab:loss_comp}, this new loss $\mathcal{L}_s^\ast$ with 2 times of weight $\alpha_0$ gets a slightly higher performance than $\mathcal{L}_s$.

As we have mentioned in Sec \ref{sec:staged_learning}, there are different kinds of deep architecture for the sketch parsing. We present the comparison of our staged learning strategy and other methods under two types of settings (w/o and w/ improvements) in Table \ref{tab:staged_comp}. It can be seen that the way of training independently for each sketch category has the worst performance under both settings. Taking advantages of the half-shared network, the models trained under the super and full branch architectures get higher average IOU scores, which show the importance of information sharing. With the proposed staged learning strategy, the parsing model achieves the best score among them. Compared to the super branch architecture, we gain 1.24\% performance improvement when used alone in the model training. As shown by the rows below ``w/ improvements'', after adopting the proposed erasing augmentation, homogeneous transformation, and soft-weighted loss, the benefit brought by staged learning is diluted to 0.68\%.

\begin{figure}[h]
    \centering
    \centerline{\includegraphics[width=1.0\linewidth]{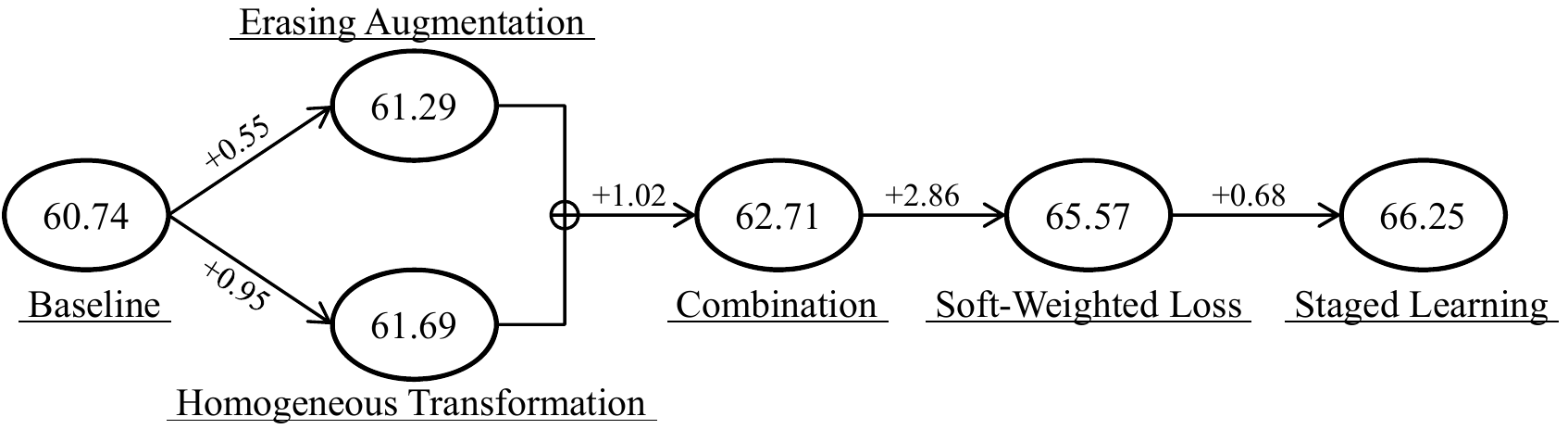}}
    \caption{Contributions of each component for our final deep sematic sketch parsing (DeepSSP) model. The baseline refers to the model trained with the augmentation methods of rotation and mirroring, the standard cross entropy loss, and the super branch architecture.}
    \label{fig:improv}
\end{figure}

Furthermore, we evaluate the contributions of each component for our final deep sematic sketch parsing (DeepSSP) model, as shown in Fig. \ref{fig:improv}. The baseline model is trained with the augmentation methods of rotation and mirroring, the standard cross entropy loss, and the super branch architecture. The results show that the proposed components improve the performance with varying magnitudes, which proves the practical value of our methods.

\subsection{Homogeneous Transformation for SBIR}
The domain adaptation is also a common problem in the field of sketch-based image retrieval (SBIR). To demonstrate the practical value of the proposed homogeneous transformation (HT) method, we integrate it into the training pipeline of existing SBIR networks and evaluate their performance on the QMUL FG-SBIR dataset \cite{song2017deep} \cite{yu2016sketch}.

The QMUL FG-SBIR dataset is constructed for the task of fine-grained instance-level SBIR. It includes three sub-datasets: shoe, chair, and handbag, in which there are 419, 297, and 568 sketch-photo pairs, respectively. The standard split of training and testing is provided by the authors and also adopted in our experiments.

We select two cutting-edge methods named Triplet SN \cite{yu2016sketch} and DSSA \cite{song2017deep} as our baseline models. Considering that both methods take triplets as the input of their networks, we apply the proposed HT method to create new training sketches as the anchor samples, which preserve the same number of triplets for the model training. Following the works of Triplet SN \cite{yu2016sketch} and DSSA \cite{song2017deep}, we use the same experimental settings and take the top K accuracy (acc.@K) as the evaluation metric,
\begin{equation}
	\begin{split}
        acc.@K = \frac {\sum_{i=1}^{n_t} r_i} {n_q}
    \end{split}
    \label{eq:score_i}
\end{equation}
in which $n_q$ is the number of query sketches in the test set of QMUL FG-SBIR dataset. When the real image paired to the query sketch is included in the top $K$ returned images, $r_i=1$, otherwise $r_i=0$.

%The comparative results against baselines on the QMUL FG-SBIR dataset (acc.@1) are shown in Table \ref{tab:sbir}. We can see that there are significant performance improvements for both baseline networks when integrated with the proposed HT method. Furthermore, we also show the results of top K accuracies (acc.@$K=1,\dots,10$) between DSSA \cite{song2017deep} and with the proposed HT method in Table \ref{tab:topkdssa}. It can be observed that the models trained with our HT method mostly perform better than the baseline methods. The experimental results demonstrate the effectiveness of the proposed HT method for fine-grained instance-level SBIR.

In Table \ref{tab:topkdssa}, we show the top K accuracies (acc.@$K=1,\dots,10$) after implementing the proposed HT method on DSSA \cite{song2017deep}. It can be observed that with the help of our HT method, DSSA exhibits better performance. Furthermore, we also present the comparative results against baselines on the QMUL FG-SBIR dataset (acc.@1) in Table \ref{tab:sbir}. It can be seen that there are significant performance improvements for both baseline networks when integrated with the proposed HT method. The experimental results demonstrate the effectiveness of the proposed HT method for fine-grained instance-level SBIR.

\begin{table}[htbp]
\centering
\caption{Comparative results against baselines on the QMUL FG-SBIR dataset (acc.@1).}
\begin{tabular}{clccc}
\toprule
{} & {} & {~~~Ori~~~}  & {~~~HT~~~} & {~~Improvement~~}\\
\midrule
\multirow{2}{*}{Shoe} & {~Triplet SN \cite{yu2016sketch}~} & 52.17 & \textbf{58.26} & +6.09\\
                {} & {~DSSA \cite{song2017deep}~} & 58.26 & \textbf{66.09} & +7.83\\
\midrule
\multirow{2}{*}{Chair} & {~Triplet SN \cite{yu2016sketch}~} & 72.16  & \textbf{82.47} & +10.31\\
                {} & {~DSSA \cite{song2017deep}~} & 79.38 & \textbf{85.57} & +6.19\\
\midrule
\multirow{2}{*}{Handbag} & {~Triplet SN \cite{yu2016sketch}~} & 39.88 & \textbf{42.86} & +2.98\\
                {} & {~DSSA \cite{song2017deep}~} & 48.21 & \textbf{50.60} & +2.39\\
\bottomrule
\end{tabular}
\begin{tablenotes}
\item $^{\ast}$ ``Ori'' means training on the original dataset without HT.
\end{tablenotes}
\label{tab:sbir}
\end{table}

%\begin{table}[h]
%\centering
%\caption{Comparative results against baselines on the QMUL FG-SBIR dataset (acc.@1).}
%\begin{tabular}{clcccc}
%\toprule
%{} & {} & {~~Ori~~} & {~~no\_fc~~} & {~~HT~~} & {~~HT\_no\_fc~~}\\
%\midrule
%\multirow{2}{*}{Shoe} & {Triplet SN \cite{yu2016sketch}} & 52.17 & 53.04 & 58.26 & 53.04\\
%                {} & {DSSA \cite{song2017deep}} & 58.26 & 57.39 & 66.09 & 56.52\\
%\midrule
%\multirow{2}{*}{Chair} & {Triplet SN \cite{yu2016sketch}} & 72.16 & 80.41 & 82.47 & 78.35\\
%                {} & {DSSA \cite{song2017deep}} & 79.38 & 85.57 & 85.57 & 86.60\\
%\midrule
%\multirow{2}{*}{Handbag} & {Triplet SN \cite{yu2016sketch}} & 39.88 & 42.86 & 42.86 & 44.64\\
%                {} & {DSSA \cite{song2017deep}} & 48.21 & 54.17 & 50.60 & 52.38\\
%\bottomrule
%\end{tabular}
%\label{tab:sbir}
%\end{table}

\begin{table*}[t]
\centering
\caption{Comparison of different methods on the SketchParse dataset.}
\begin{tabular}{lcccccccccccc}
\toprule
{} & {~~cow~~} & {~~horse~~} & {~~cat~~} & {~~dog~~} & {~~sheep~~} & {~~bus~~} & {~~car~~} & {bicycle} & {motorbike} & {airplane} & {~~bird~~} & {~~average~~}\\
\midrule
{FCN \cite{long2015fully}~~} & 61.76 & 66.69 & 63.45 & 68.37 & 65.70 & 63.94 & 56.13 & 52.36 & 46.32 & 51.65 & 35.16 & 57.05\\
{Deeplab \cite{chen2017rethinking}~~} & 65.34 & 68.11 & 64.49 & 68.33 & 66.88 & 63.42 & 57.09 & 56.95 & 47.82 & 54.23 & 36.68 & 58.77\\
{Baseline~~} & 66.01 & 67.77 & 66.37 & 67.41 & 67.37 & 65.80 & 63.15 & 59.15 & 50.43 & 52.95 & 44.57 & 60.74\\
{MM 17' \cite{sarvadevabhatla2017sketchparse}~~} & 68.78 & 69.35 & 69.60 & 71.18 & 70.81 & \textbf{68.00} & 67.35 & 62.66 & 55.04 & 57.34 & 50.89 & 64.45\\
{Our method~~} & \textbf{70.42} & \textbf{72.81} & \textbf{69.94} & \textbf{72.57} & \textbf{72.02} & 67.88 & \textbf{70.88} & \textbf{63.30} & \textbf{55.69} & \textbf{59.96} & \textbf{54.38} & \textbf{66.25}\\
\bottomrule
\end{tabular}
\label{tab:comp}
\end{table*}

\subsection{Comparative Results}
\begin{figure}[htbp]
    \centering
    \centerline{\includegraphics[width=1.0\linewidth]{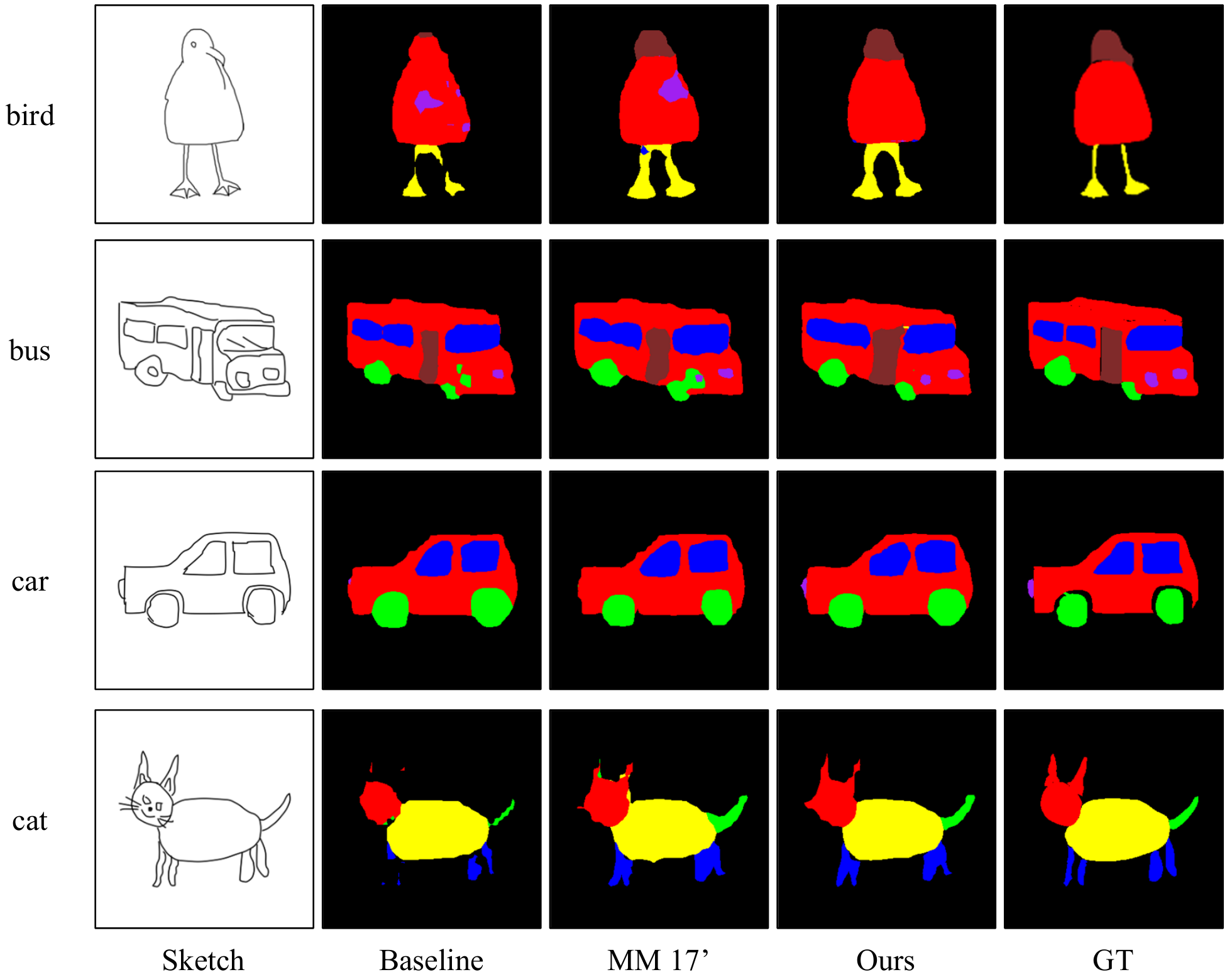}}
    \caption{Parsing results of different methods. From left to right: input freehand sketches, outputs of three methods (i.e., the baseline, MM 17', and our model), ground truth annotations.}
    \label{fig:result_comp}
\end{figure}

Table \ref{tab:comp} shows the comparison of our method against FCN \cite{long2015fully}, Deeplab \cite{chen2017rethinking}, baseline, and the work of Sarvadevabhatla et al. \cite{sarvadevabhatla2017sketchparse} (noted as MM 17'). Both FCN and Deeplab models are trained independently for each sketch category. In comparison, the baseline model, which is trained with the rotation and mirroring augmentation, standard cross entropy loss, and super branch architecture, gets a higher average IOU score (60.74\%). With the proposed DeepSSP framework, our method achieves 5.51\% higher performance than the baseline. Especially, the performance on category ``bird'' is improved by nearly 10 points. Our method also beats MM 17' in 10 of 11 sketch categories. For categories ``car'' and ``bird'', we achieve 3.53\% and 3.49\% performance improvements. By integrating the proposed methods together, our final model becomes the new state-of-the-art on the task of part-level semantic sketch parsing. Some examples of the parsing results are shown in Fig. \ref{fig:result_comp}. Compared to other methods, our predictions look more accurate in parts like the wheel of the bus and the headlight of the car, which shows the superiority of our methods.

\begin{figure}[t]
    \centering
    \centerline{\includegraphics[width=1.0\linewidth]{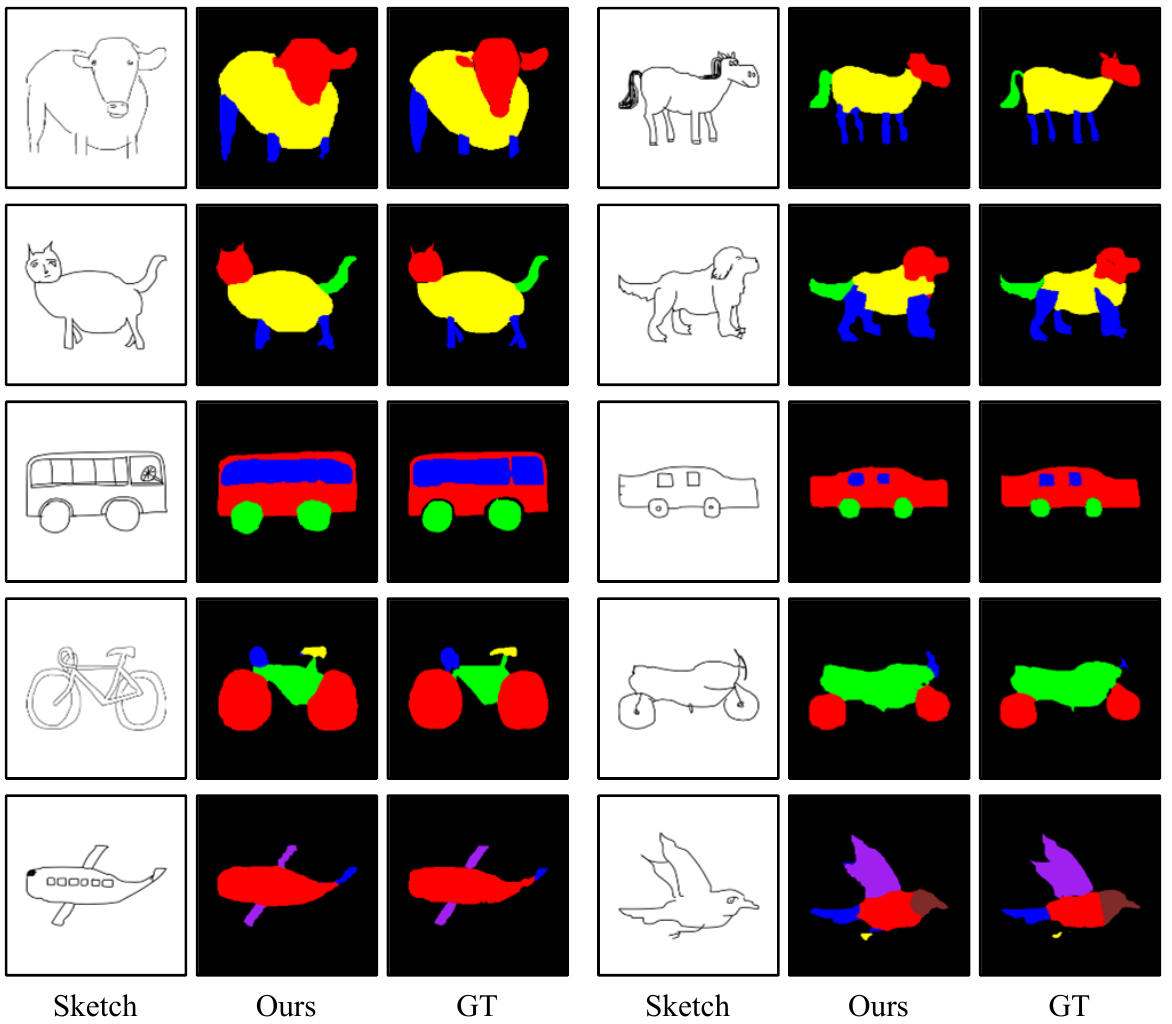}}
    \caption{Illustrations of results output by our final DeepSSP model.}
    \label{fig:results_ssp}
\end{figure}

\subsection{Discussion}
\begin{figure}[h]
    \centering
    \centerline{\includegraphics[width=1.0\linewidth]{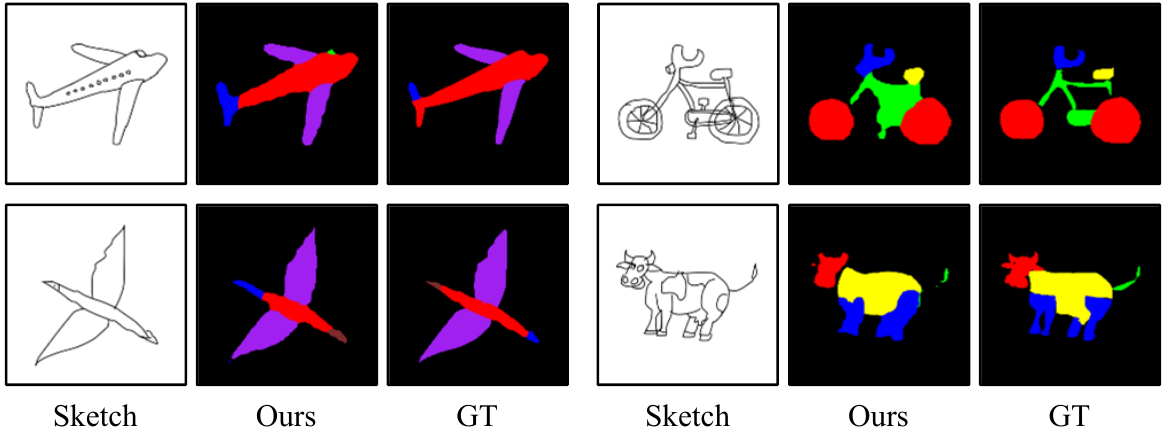}}
    \caption{Some failure cases of our method, taken from categories of airplane, bicycle, bird, and cow.}
    \label{fig:failure}
\end{figure}

Fig. \ref{fig:results_ssp} shows some representative results output by our final DeepSSP model. As shown in the figure, our DeepSSP works well on different categories of freehand sketches, which demonstrate the effectiveness of our method for the task of part-level semantic sketch parsing. Furthermore, some failure cases are shown in Fig. \ref{fig:failure}. The first figure shows our result of the category ``airplane'', which has the wrong prediction at the position of the cockpit window. The failure for the second result of the category ``bicycle'' lies in the bicycle frame, which is normally labeled as a hollow part. However, sometimes the bicycle frame has a filled annotation as shown in Fig. \ref{fig:results_ssp}. These two kinds of failures are mostly caused by the ambiguous part-level annotations. The third case for the category ``bird'' mixes up the positions of the part class ``head'' and ``tail''. If the extra information on the spatial relationship between different part classes is provided, it could output a better parsing result. The last one from the category ``cow'' misclassified the background pixels between legs. A finer dense prediction is required to solve this problem, which is left to the future work.

From Table \ref{tab:comp}, we can see that the performance of our method varies across different sketch categories. For example, there is 18.43\% gap between the category ``horse'' and ``bird''. To give a quantitative analysis for these sketch categories with worse performance, we present their detailed IOU scores of part classes in Table \ref{tab:iou_class}. The results show that the loss of these categories is mainly due to poor performance on part classes with a smaller size than others, such as the ``engine'' of ``airplane'' and ``handlebar'' of ``bicycle''. For categories ``bicycle'' and ``motorbike'', there is an unexpected difference in the part class ``seat'' (50.19\% vs 8.8\%). By analyzing sketches from two categories, as illustrated in the fourth row of Fig. \ref{fig:results_ssp}, we found that the ``seat'' is a common part in ``bicycle'', but not usually in ``motorbike''. It makes the prediction of ``seat'' of ``motorbike'' more difficult. To improve the parsing quality of these sketch categories, we are exploring new insights based on balance learning.

\begin{table}[htbp]
\centering
\caption{Detailed IOU scores of part classes for sketch categories with worse performance.}
\begin{tabular}{cccccc}
\toprule
\multirow{2}{*}{airplane} & {body} & {window} & {tail} & {engine} & {wing}\\
{} & 68.36 & 37.01 & 42.34 & 16.13 & 46.91\\
\midrule
\multirow{2}{*}{bird} & {body} & {leg} & {tail}& {head} & {wing} \\
{} & 53.60 & 32.00 & 30.32 & 47.96 & 44.66\\
\bottomrule
\toprule
{} & {wheel} & {seat} & {handlebar} & {body}\\
\midrule
{bicycle} & 89.37 & 50.19 & 22.08 & 59.09\\
{motorbike} & 82.68 & 8.88 & 21.30 & 60.00\\
\bottomrule
\end{tabular}
\label{tab:iou_class}
\end{table}

\section{Conclusion}
\label{sec:conclusion}
Our novel DeepSSP framework re-purposes the network designed for real image segmentation to the task of part-level semantic freehand sketch parsing by integrating the homogeneous transformation, soft-weighted loss, and staged learning. We propose the homogeneous transformation to solve the problem of the semantic gap between the domains of real images and freehand sketches. To avoid the dilemma of ambiguous label boundary and class imbalance, we reshape the standard cross entropy loss to the soft-weighted loss for better guidance for the model training. Furthermore, we present a staged learning strategy that takes advantages of the shared information across categories and the specific characteristic of each sketch class. Extensive experimental results prove the practical value of our method and show that our final DeepSSP achieves the state-of-the-art on the public SketchParse dataset.

\ifCLASSOPTIONcaptionsoff
  \newpage
\fi


\bibliographystyle{IEEEtran}
\bibliography{refs}
\vspace{-30pt}

\begin{IEEEbiography}[{\includegraphics[width=1in,height=1.25in,clip,keepaspectratio]{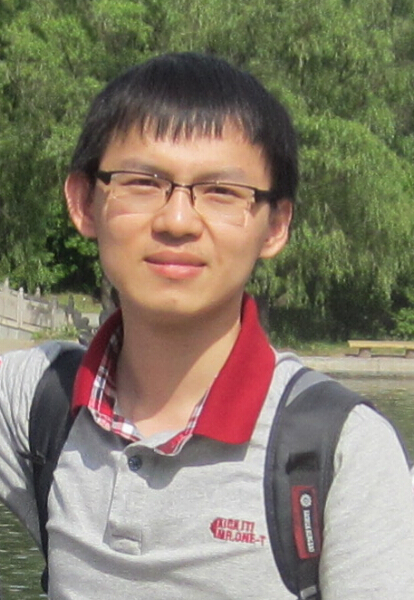}}]{Ying Zheng} received the Ph.D. and M.S. degrees in computer science from the School of Computer Science and Technology, Harbin Institute of Technology, Harbin, China, in 2019 and 2014. He is currently an associate researcher of Zhejiang Lab. He was a visiting student with the Australian National University, ACT, Australia. His research interests include computer vision and deep learning, especially focusing on understanding of freehand sketches.
\end{IEEEbiography}

\vspace{-5pt}

\begin{IEEEbiography}[{\includegraphics[width=1in,height=1.25in,clip,keepaspectratio]{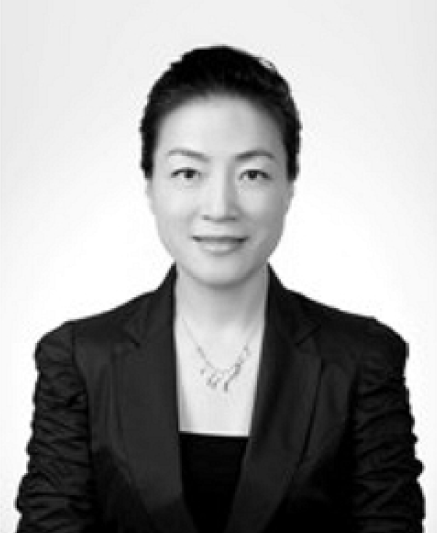}}]{Hongxun Yao} received the B.S. and M.S. degrees in computer science from the Harbin Shipbuilding Engineering Institute, Harbin, China, in 1987 and in 1990, respectively, and received Ph.D. degree in computer science from Harbin Institute of Technology in 2003. Currently, she is a professor with the School of Computer Science and Technology, Harbin Institute of Technology. Her research interests include computer vision, pattern recognition, multimedia computing, human-computer interaction technology. She has 6 books and over 200 scientific papers published, and won both the honor title of ``the new century excellent talent'' in China and ``enjoy special government allowances expert'' in Heilongjiang Province, China.
\end{IEEEbiography}

\vspace{-5pt}

\begin{IEEEbiography}[{\includegraphics[width=1in,height=1.25in,clip,keepaspectratio]{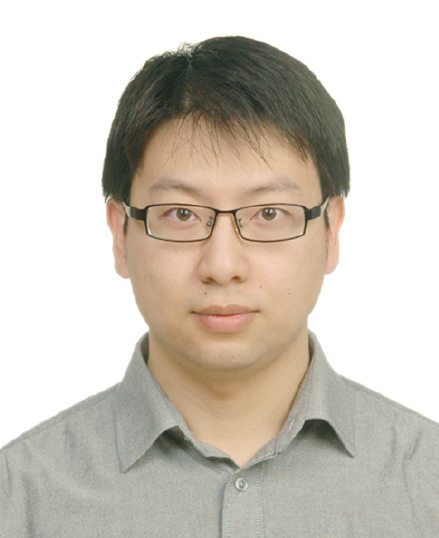}}]{Xiaoshuai Sun} received the B.S. degree in Computer Science from Harbin Engineering University in 2007. He received the M.S. and Ph.D. degree in Computer Science and Technology from Harbin Institute of Technology in 2009 and 2015 respectively. He is currently an associate professor of Xiamen University. He was a Research Intern with Microsoft Research Asia (2012-2013) and also a winner of Microsoft Research Asia Fellowship in 2011. He invented 2 patents and authored over 60 journal and conference papers in IEEE Transactions on Image Processing, Pattern Recognition, ACM Multimedia, and IEEE CVPR.
\end{IEEEbiography}


\vfill


\end{document}